\documentclass[10pt,twoside]{article} 

\usepackage{arxiv_custom}

\usepackage[T1]{fontenc}
\usepackage{hyperref}
\usepackage{url}
\usepackage{booktabs}
\usepackage{amsfonts}
\usepackage{amsmath}
\usepackage{nicefrac}
\usepackage{microtype}
\usepackage{graphicx}
\usepackage{multirow}
\usepackage{caption}
\usepackage{subcaption}
\usepackage{placeins}
\usepackage{bm}
\usepackage{makecell}
\usepackage{pifont}
\usepackage{tikz}
\usepackage{xcolor}
\usepackage{cleveref}
\usepackage{tabularx}

\newcommand{\G}{\mathcal{G}}
\newcommand{\V}{\mathcal{V}}
\newcommand{\E}{\mathcal{E}}
\newcommand{\z}{\bm{z}}
\newcommand{\Z}{\bm{Z}}
\newcommand{\Zp}{\bm{Z'}}
\newcommand{\yhat}{\hat{\bm{y}}}

\newcommand{\Sm}{\bm{S}}
\newcommand{\Smp}{\bm{S'}}

\newcommand{\Out}{\bm{O}}
\newcommand{\Outp}{\bm{O'}}
\newcommand{\Nat}{\mathbb{N}}
\newcommand{\Real}{\mathbb{R}}
\newcommand{\pyg}{\texttt{PyTorch Geometric}}
\newcommand{\resi}{\texttt{ReSi}}

\newcommand{\subheader}[1]{\noindent\textbf{#1}}
\newcommand{\captionheader}[1]{\textbf{\textit{#1}}}

\newcommand{\bbone}{\text{\usefont{U}{bbold}{m}{n}1}}
\MakeRobust{\bbone}

\newcommand{\ols}[1]{\mskip.5\thinmuskip\overline{\mskip-.5\thinmuskip {#1} \mskip-.5\thinmuskip}\mskip.5\thinmuskip}

\colorlet{revisioncolor}{black}

\newcommand{\new}[1]{\textcolor{revisioncolor}{#1}}

\newenvironment{revisionblock}
{%
	\par
	\begingroup
	\color{revisioncolor}%
	\ignorespaces
}
{%
	\par
	\endgroup
	\ignorespacesafterend
}

\newcolumntype{L}[1]{%
	>{\raggedright\arraybackslash}m{#1}%
}

\newcolumntype{Y}{%
	>{\centering\arraybackslash}X%
}
\newcommand{\methodlabel}[1]{%
	\raisebox{3ex}{#1}%
}

\definecolor{patternteal}{RGB}{0,135,120}
\definecolor{heterogray}{RGB}{100,105,110}
\definecolor{rulegray}{RGB}{195,202,210}

\newcommand{\glyphframe}[1]{%
	\begin{tikzpicture}[x=1.05cm,y=0.82cm,baseline=-0.23cm]
		\path[use as bounding box] (0,0) rectangle (1,1);
		\draw[rulegray,line width=0.35pt] (0,.50) -- (1,.50);
		#1
		\draw[black,line width=0.45pt,line cap=round]
		(0,1) -- (0,0) -- (1,0);
	\end{tikzpicture}%
}

\newcommand{\trendglyph}[1]{%
	\glyphframe{%
		\draw[patternteal,line width=1.05pt,line cap=round,line join=round]
		plot[smooth,tension=.60] coordinates {#1};
	}%
}

\newcommand{\logglyphframe}[1]{%
	\begin{tikzpicture}[x=1.05cm,y=0.82cm,baseline=-0.23cm]
		\path[use as bounding box] (0,0) rectangle (1,1);
		\draw[rulegray,densely dashed,line width=0.40pt] (0,.25) -- (1,.25);
		\draw[rulegray,densely dashed,line width=0.40pt] (0,.50) -- (1,.50);
		\draw[rulegray,densely dashed,line width=0.40pt] (0,.75) -- (1,.75);
		\draw[rulegray,densely dashed,line width=0.40pt] (0,.99) -- (1,.99);
		#1
		\draw[black,line width=0.45pt,line cap=round]
		(0,1) -- (0,0) -- (1,0);
	\end{tikzpicture}%
}

\newcommand{\partialglyph}[1]{%
	\glyphframe{%
		\draw[patternteal,dash pattern=on 3.2pt off 2.1pt,line width=1.05pt,
		line cap=round,line join=round]
		plot[smooth,tension=.60] coordinates {#1};
	}%
}

\newcommand{\heteroglyph}[3]{%
	\glyphframe{%
		\draw[heterogray,line width=.80pt,line cap=round,line join=round]
		plot[smooth,tension=.58] coordinates
		{(0,#1) (.24,#1) (.50,#2) (.76,#3) (1,#3)};
		\draw[heterogray,line width=.80pt,line cap=round,line join=round]
		plot[smooth,tension=.58] coordinates
		{(0,#3) (.24,#3) (.50,#2) (.76,#1) (1,#1)};
	}%
}

\newcommand{\heterologglyph}[3]{%
	\logglyphframe{%
		\draw[heterogray,line width=.80pt,line cap=round,line join=round]
		plot[smooth,tension=.58] coordinates
		{(0,#1) (.24,#1) (.50,#2) (.76,#3) (1,#3)};
		\draw[heterogray,line width=.80pt,line cap=round,line join=round]
		plot[smooth,tension=.58] coordinates
		{(0,#3) (.24,#3) (.50,#2) (.76,#1) (1,#1)};
	}%
}

\newcommand{\cell}[2]{%
	\shortstack{#1\\[-2.4mm]{\tiny #2}}%
}

\newcommand{\partialcell}[2]{%
	\shortstack{#1\\[-2.4mm]{\tiny #2}}%
}

\newcommand{\heterocell}[1]{%
	\shortstack{#1\\[-2.4mm]{\tiny\phantom{0/0}}}%
}

\DeclareRobustCommand{\legendtealline}{%
	\kern-0.1em%
	\tikz[baseline=-0.65ex]{%
		\draw[
		teal,
		line width=1.05pt,
		] (0,0) -- (5mm,0);
	}%
	\kern-0.1em%
}
\DeclareRobustCommand{\legendtealdashedline}{%
	\leavevmode
	\kern-0.1em%
	\tikz[baseline=-0.65ex]{%
		\draw[
		teal,
		line width=1.05pt,
		dash pattern=on 4pt off 1.2pt
		] (0,0) -- (5.2mm,0);
	}%
	\kern-0.12em%
}
\newcommand{\legendgrayx}{\kern-0.12em \textcolor{heterogray}{$\boldsymbol{\times}$}\kern-0.12em}

\title{The Impact of Dimensionality\\ on the Stability of Node Embeddings}

\author{
    Tobias Schumacher \\
    University of Mannheim, \\
    RWTH Aachen University \\
  \texttt{tobias.schumacher@uni-mannheim.de} 
   \And
    Simon Reichelt \\
    University of Mannheim \\
    \texttt{simon.marius.reichelt@gmail.com}
   \AND
    Markus Strohmaier \\
    University of Mannheim,\\
    GESIS - Leibniz Institute for the Social Sciences, and\\
    Complexity Science Hub \\
    \texttt{markus.strohmaier@uni-mannheim.de}
}

\begin{document}

\maketitle

\begin{abstract}
Previous work has shown that node embedding methods can produce different representations and downstream predictions across repeated training runs, even when trained on the same data with identical hyperparameters. 
However, the role of embedding dimensionality in this instability remains poorly understood.
In this work, we systematically analyze how embedding dimensionality affects the stability of embeddings from five widely used node embedding methods: ASNE, DGI, GraphSAGE, node2vec, and VERSE. We evaluate stability from both representational and functional perspectives across a broad range of dimensions, datasets, and repeated training runs, and relate the resulting stability patterns to predictive performance.
Our results show that dimensionality can substantially affect embedding stability, although the observed effects depend strongly on the embedding method and stability notion considered. While node2vec and ASNE generally became more stable at higher dimensions, GraphSAGE and VERSE often exhibited non-monotonic behavior or decreasing stability. We further find that dimensions associated with high stability do not necessarily coincide with those yielding the strongest downstream performance. Overall, our findings demonstrate that embedding dimensionality can have a substantial impact on the stability of node embeddings and downstream predictions.
\end{abstract}

\section{Introduction}

Node embedding methods provide a well-established approach for representing network data. By mapping nodes into a continuous vector space, they facilitate the analysis of complex networks, including tasks such as node classification, link prediction, community discovery, and visualization~\citep{goyal_graph_2018, chen_graph_2020}. A fundamental design choice in this process is the embedding dimensionality, which determines the size of the learned representations. Prior work has shown that dimensionality can substantially affect downstream performance~\citep{goyal_graph_2018}, motivating the development of methods for dimensionality selection~\citep{gu_principled_2021,luo_graph_2021, yuki_dimensionality_2023} and intrinsic dimensionality estimation~\citep{nakis_how_2025}. 
Beyond predictive performance, the stability of node embeddings has recently attracted growing interest. Repeated runs of the same embedding algorithm can produce different representations and downstream predictions, even when trained on the same graph with identical hyperparameters~\citep{schumacher_effects_2021, wang_understanding_2022}. Such variability can complicate the interpretation of learned representations and affect the consistency of results obtained from embedding-based network analyses. 

However, the interplay between embedding dimensionality and stability  remains largely unexplored. While prior work has occasionally considered dimensionality when analyzing stability~\citep{wang_understanding_2022, gu_principled_2021}, it has typically been treated as one factor among many or examined only in limited settings.
In this work, we address this gap by pursuing three objectives: (i) to characterize how embedding dimensionality affects stability, (ii) to analyze the relationship between stability and downstream performance across different embedding dimensions, and (iii) to investigate how graph size and density influence the effect of dimensionality on stability.

To achieve these objectives, we analyze five widely used node embedding methods---ASNE~\citep{liao_attributed_2018}, DGI~\citep{velickovic_deep_2019}, GraphSAGE~\citep{hamilton_inductive_2017}, node2vec~\citep{grover_node2vec_2016}, and VERSE~\citep{tsitsulin_verse_2018}---across dimensions ranging from 4 to 4096. We evaluate stability from both representational and functional perspectives using multiple complementary measures. In addition to seven empirical benchmark datasets, we consider synthetic graphs to investigate how graph size and density influence dimensionality-dependent stability. Overall, we trained and analyzed more than 150,000 embeddings, enabling a comprehensive analysis of how dimensionality affects stability across embedding methods and network settings.

\section{Background}\label{sec:background}
Before describing our experimental framework, we briefly provide background on node embeddings and their stability and summarize related work on the topic.

\subsection{Preliminaries}\label{sec:prelim}
We briefly introduce the core concepts and notation required for our analyses.

\subheader{Graphs.}
We define a graph $\G$ as a tuple $\G=(\V,\E)$, where $\V=(v_1,\dots,v_N)$ denotes the set of vertices and $\E=(e_1,\dots,e_M)$ the set of edges.
Each edge $e=(v_i,v_j)\in\E$ corresponds to a relation between two nodes $v_i,v_j\in\V$.
In addition, nodes and edges may be associated with features---for simplicity, we omit corresponding notation.

\subheader{Node Embeddings.}
Node embedding algorithms map nodes of a graph to (ideally) low-dimensional vectors that encode structural information about the graph and, potentially, node or edge features, such that these vectors can be used for downstream tasks.
Formally, a node embedding $\varphi:\V\rightarrow\Real^D$ maps each node $v_i\in\V$ to an embedding vector $\z_i=\varphi(v_i)$ of dimension $D\in\Nat$.
By stacking these vectors, one obtains an embedding matrix $\Z=[\z_1,\dots,\z_N]^T\in\Real^{N\times D}$.

\subheader{Downstream Tasks.}
After embeddings have been generated, they can be used for downstream tasks such as node classification or link prediction.

In node classification, each node $v_i\in\V$ is associated with a label $y_i\in\mathcal{Y}=\{1,\dots,C\}$.
The goal is to train a classifier $c_{\operatorname{NC}}:\Real^D\rightarrow\mathcal{Y}$ on embeddings of nodes with known labels and predict labels for nodes whose labels are unknown.
We denote the output of the classifier for node $v_i$ as $\Out_i=c_{\operatorname{NC}}(\z_i)\in\Real^C$, where $\Out_i$ contains class-wise prediction scores, and let  $\yhat_i:=\arg\max\Out_i$ for the predicted class for node $v_i$.
For an ordered set of $n\leq N$ classified nodes, we correspondingly consider the output matrix $\Out\in\Real^{n\times C}$ and the vector of hard predictions $\bm{y}\in\Real^n$.

In link prediction, the goal is to predict missing or future links based on the observed graph structure.
Accordingly, a classifier $c_{\operatorname{LP}}$ operates on pairs of node embeddings $(\z_i,\z_j)$ corresponding to node pairs $(v_i,v_j)$.
Since the classifier requires a single input representation, pairs of embeddings are commonly combined through operations such as concatenation, averaging, weighted differences, or the Hadamard product~\citep{grover_node2vec_2016, tsitsulin_verse_2018}.
For simplicity, we overload notation and also denote the resulting outputs for an ordered set of $m\leq \binom{N}{2}$ node pairs as $\Out\in\Real^{m\times 2}$.
In practice, link prediction is formulated as a binary classification task over existing and non-existing edges. Since real-world graphs are typically sparse, considering all possible node pairs would yield highly imbalanced training data and be computationally infeasible. Thus, negative edges are usually sampled during training.

\subheader{Stability of Node Embeddings.}
Due to stochasticity in training, repeated runs of the same node embedding algorithm can yield different embeddings under identical graph and hyperparameter settings.
In this context, stability refers to the extent to which embeddings and their induced predictions remain consistent across such runs.
We study the stability of node embeddings from two complementary perspectives based on the notions of \textit{representational stability} and \textit{functional stability}.

\emph{Representational stability} describes the extent to which embeddings $\Z$ produced across different runs are similar to each other. \emph{Functional stability}, by contrast, concerns the consistency of downstream predictions $\Out$ obtained from embeddings. Correspondingly, representational and functional stability are quantified using \emph{representational similarity measures} and \emph{functional similarity measures}, respectively. 
\new{Considering both perspectives allows us to assess whether variability in embeddings is reflected in the derived downstream predictions.
	These two notions need not coincide:}
embeddings that differ substantially in representation space may nevertheless yield similar downstream predictions, and vice versa~\citep{klabunde_similarity_2025}. We discuss the measures used in this study in Section \ref{sec:evaluation} and refer to the survey by \citet{klabunde_similarity_2025} for a broader overview of representational and functional similarity.

\subsection{Related Work}
It is well established that different random seeds can lead to different outcomes in machine learning models~\citep{picard_torchmanual_seed3407_2023, bouthillier_accounting_2021}. 
This issue has also been studied for learned embeddings, most prominently in the word and node embedding domains. 
Since several node embedding methods and analyses in graph representation learning build on ideas from word embeddings, we first review related work on dimensionality and stability in word embeddings. 
We then discuss stability and dimensionality in node embeddings before focusing on prior work that explicitly connects these two aspects.

\subheader{Dimensionality and Stability of Word Embeddings.}
Choosing an appropriate embedding dimension can strongly influence the performance of word embeddings~\citep{lai_how_2016, chiny_effect_2023}. Consequently, several works have investigated principled approaches for selection embedding dimension. 
Most notably, \citet{yin_dimensionality_2018} introduced the \emph{PIP loss} as a means to identify optimal dimensionality, and identified a bias-variance trade-off for matrix factorization-based embeddings.
Building on this, \citet{wang_single_2019} proposed a PCA-based approach to directly estimate suitable embedding dimension.

The instability of word embeddings was first highlighted by \citet{hellrich_bad_2016}, who observed substantial variation in nearest neighbors despite identical training configurations differing only in the choice of random seed. Subsequent work confirmed and further analyzed this phenomenon~\citep{antoniak_evaluating_2018, borah_are_2021, dridi_knn_2018, chugh_stability_2018, leszczynski_understanding_2020}. In particular, \citet{wendlandt_factors_2018} and \citet{pierrejean_investigating_2019} investigated factors contributing to embedding instability, while \citet{dridi_knn_2018} proposed considering stability in hyperparameter tuning.

\subheader{Node Embedding Stability.}
Node embedding stability has most prominently been analyzed by \citet{schumacher_effects_2021} and \citet{wang_understanding_2022}. \citet{schumacher_effects_2021} investigated the stability of embeddings from both a geometric perspective and a downstream perspective. Considering HOPE~\citep{ou_asymmetric_2016}, LINE~\citep{tang_line_2015}, SDNE~\citep{wang_structural_2016}, node2vec~\citep{grover_node2vec_2016}, and GraphSAGE~\citep{hamilton_inductive_2017}, they found substantial differences in embedding spaces for most methods, while aggregate downstream performance remained comparatively stable despite variation in individual predictions. Similar findings regarding geometric instability were reported by \citet{wang_understanding_2022} for LINE, SDNE, node2vec, DeepWalk~\citep{perozzi_deepwalk_2014}, struc2vec~\citep{ribeiro_struc2vec_2017}, and DNGR~\citep{cao_deep_2016}. They also conducted a regression analysis identifying graph properties, such as network size, density, and assortativity, and algorithmic parameters, such as random walk length and number, as significant factors influencing stability.

Additional work has analyzed stability in graph neural networks and related embedding models. \citet{klabunde_prediction_2022} observed that while aggregate predictive performance remained largely stable across runs, individual predictions still differed substantially. Complementing this work, \citet{shi_geometric_2023} proposed the \emph{graph Gram index} as a measure of geometric stability in graph neural networks. Finally, \citet{meroue_link_2026} recently analyzed stability in knowledge graph embeddings, likewise reporting substantial variation in both embedding spaces and individual predictions.

\subheader{Dimensionality of Node Embeddings.}
In the context of node embeddings, it has been well established that embedding dimensionality can strongly affect downstream performance. Prior work has shown that increasing dimensionality typically improves performance up to a threshold, after which performance either plateaus or deteriorates~\citep{goyal_graph_2018, chen_graph_2020}. Consequently, several approaches have been proposed to determine embedding dimensionality in a principled manner.

One line of work has investigated intrinsic graph dimensionality, pointing out that the number of dimensions required to reconstruct even larger real-world networks is typically very low~\citep{bonato_dimensionality_2014, almagro_detecting_2022, nakis_how_2025}. Yet, these works are mostly aimed at the compression of complex networks rather than training of machine learning-oriented embeddings.

A separate line of work has focused on selecting embedding dimensions directly.
Building on related approaches for word embeddings, \citet{gu_principled_2021} adapted the \emph{PIP loss} to node embeddings and used it to estimate an embedding dimension that preserves structural information while minimizing information loss.
Similarly, \citet{luo_graph_2021} and \citet{xu_optimal_2023} proposed entropy-based approaches for selecting embedding dimension for graph neural networks. Additional work has also built dimension selection directly into embedding models and optimization procedures~\citep{dong_deep_2024, yuki_dimensionality_2023}.

\subheader{Relationship Between Dimensionality and Stability.}
In the word embedding domain, several works analyzing stability considered embedding dimensionality as a potential factor. 
\citet{leszczynski_understanding_2020} found that downstream stability of several embedding methods improved with increasing dimension up to a plateau, which was reached between dimensions 200 and 400. 
Similar trends were observed for nearest-neighbor stability in embedding space, which often plateaued around $D=300$~\citep{dridi_knn_2018,chugh_stability_2018,borah_are_2021}.

By contrast, the relationship between dimensionality and stability has received considerably less attention in the node embedding domain. 
In their regression analysis of factors influencing stability, \citet{wang_understanding_2022} included embedding dimension as a linear predictor and found a statistically significant, moderately sized effect on embedding stability. 
However, dimensionality was not analyzed systematically, and potential non-linear relationships between dimensionality and stability remained unexplored. 
In a brief side analysis, \citet{gu_principled_2021} also evaluated how node2vec stability varied across increasing dimensions, observing that embedding variance initially decreased before slightly increasing again at higher dimensions. 
These findings suggest that dimensionality may affect stability in non-trivial ways, but existing evidence remains limited to isolated analyses. We therefore provide a systematic investigation across multiple methods, datasets, and graph properties.

\section{Experiments}

The objective of this study is to examine the relationship between node embedding dimensionality, stability, and downstream performance across a range of embedding methods and datasets. In addition, we investigate how graph characteristics influence these relationships by systematically varying structural properties such as graph size and density in synthetic networks. To this end, we train embeddings across multiple dimensions and evaluate representational stability across all settings, while functional stability and downstream performance are assessed on real-world datasets. This section outlines the experimental methodology used in our analysis. All code used for our experiments is available on \url{https://github.com/dess-mannheim/dimpact}.

\subsection{Node Embedding Algorithms}
We consider the following five node embedding algorithms, aiming to cover a broad range of methodological approaches as well as sources of randomness and instability.

\begin{itemize}
	\item \textit{\textbf{Attributed Social Network Embedding (ASNE)}}~\citep{liao_attributed_2018} jointly \new{incorporates graph structure and node attributes into node representations.
		In the evaluated implementation, embeddings are learned using \emph{Doc2Vec}~\citep{le_distributed_2014}, introducing randomness through vector initialization, negative sampling, stochastic downsampling of frequent tokens, and multithreaded training.}
	
	\item 
	\textit{\textbf{Deep Graph Infomax (DGI)}}~\citep{velickovic_deep_2019} is a graph representation learning approach based on maximizing mutual information between local node representations and global graph summaries using a graph convolutional encoder~\citep{kipf_semisupervised_2017}. Randomness is introduced through weight initialization, \new{neighborhood sampling, minibatch shuffling,} and stochastic corruption of node features.

	\item 
	\textit{\textbf{GraphSAGE}}~\citep{hamilton_inductive_2017} applies a graph neural network (GNN)~\citep{scarselli_graph_2009}, which iteratively aggregates representations from sampled node neighborhoods. Randomness is introduced through neighborhood sampling, \new{negative-edge sampling, minibatch shuffling,} and weight initialization.
	
	\item 
	\textit{\textbf{node2vec}}~\citep{grover_node2vec_2016} learns node representations using random walks and a skip-gram objective inspired by \emph{word2vec}~\citep{mikolov_distributed_2013}. Sources of randomness include the generation of random walks, model initialization, \new{multithreaded optimization}, and negative sampling during optimization.
	
	\item \textit{\textbf{VERSE}}~\citep{tsitsulin_verse_2018} learns embeddings by approximating graph-based node similarity measures in the embedding space using a shallow neural architecture. Randomness is introduced through embedding initialization, \new{multithreaded optimization}, and the sampling procedures used during optimization via noise-contrastive estimation~\citep{gutmann_noisecontrastive_2010}.
\end{itemize}

\new{The sources of randomness identified above refer to the specific implementations evaluated in our experiments.
	We implemented DGI and GraphSAGE using \texttt{PyTorch Geometric}~\citep{fey_fast_2019,fey_pyg_2025}.
	For node2vec and ASNE, we used the implementations provided by \texttt{GRAPE}~\citep{cappelletti_grape_2023} and \texttt{karateclub}~\citep{rozemberczki_karate_2020}, respectively, while for VERSE, we used the reference implementation~\citep{tsitsulin_verse_2018}.} More details on the implementations are provided in Appendix~\ref{ap:dimpact_params}.

\subsection{Datasets}
To analyze the relationship between dimension, stability, and performance, we trained embeddings on a diverse set of empirical and synthetic datasets.

\subheader{Empirical Datasets.}
We evaluate our methods on a diverse set of empirical graph datasets covering citation, social, hyperlink, and biological interaction networks. An overview of all datasets and their properties is also provided in Table~\ref{tab:datasets}.

Specifically, we consider the citation networks \textit{Cora} and \textit{PubMed}~\citep{yang_revisiting_2016a}; the social networks \textit{BlogCatalog}~\citep{yang_pane_2023} and \textit{Facebook}~\citep{rozemberczki_multiscale_2021}; the hyperlink network \textit{Wikipedia}~\citep{yang_pane_2023}; the drug-drug interaction network \textit{OGBL-DDI}~\citep{hu_open_2020}; and the \textit{CoAuthor} network~\citep{tsitsulin_verse_2018} that was extracted from the Microsoft Academic Graph~\citep{sinha_overview_2015}.
We obtained Cora, PubMed, Wikipedia, BlogCatalog, and Facebook from \pyg{} \citep{fey_fast_2019,fey_pyg_2025}, OGBL-DDI from the Open Graph Benchmark~\citep{hu_open_2020}, and for CoAuthor, we used data processed by \citet{tsitsulin_verse_2018}. More details are provided in Appendix \ref{ap:datasets}.

\begin{table}[b]
	\centering
	\caption{\textbf{\textit{Overview of empirical datasets.}} 
		\emph{Features} indicates the number of node features, and
		\emph{Classes} indicates  number of target classes in the downstream task.}
	\label{tab:datasets}\smallskip
	\begin{tabular}{lcccccc}
		\toprule
		\textbf{Dataset} & \textbf{Nodes} & \textbf{Edges} & \textbf{Density} & \textbf{Features}  & \textbf{Downstream Task} & \textbf{Classes}\\
		\midrule
		Cora & 2,708 & 10,556 & 0.0029 & 1,433 & Node Classification & 7\\
		PubMed & 19,717 & 44,338 & 0.0002 & 500 & Node Classification & 3 \\ 
		BlogCatalog & 5,196 & 343,486 & 0.0254 & 8,189 & Node Classification & 6\\
		Facebook & 22,407 & 342,004 & 0.0014 & 128 & Node Classification & 4\\ 
		Wikipedia & 2,405 & 17,981 & 0.0062 & 4,973 & Node Classification & 17\\
		OGBL-DDI & 4,267 & 1,334,889 & 0.1467 & - & Link Prediction & 2 \\
		CoAuthor & 51,820 & 177,963 & 0.0001 & - & Link Prediction & 2\\
		\bottomrule
	\end{tabular}%
\end{table}

\subheader{Synthetic Datasets.}
Following \citet{schumacher_effects_2021}, we also evaluate the impact of network size and density on the stability of node embeddings using synthetic \emph{Barabasi-Albert} (BA)~\citep{barabasi_emergence_1999a} and \emph{Watts-Strogatz} (WS)~\citep{watts_collective_1998} networks.
For the WS networks, we chose a rewiring probability $p=0.1$, such that both network models would yield high clustering and the small-world property. The main difference between these models lies in the degree distribution, which follows a power law curve for the BA networks and is homogeneous for WS networks.
To analyze the effect of density, we generated networks with fixed size $N=1600$ and varying density
\begin{equation}
	d\in\{0.001, 0.002, 0.005, 0.01, 0.02, 0.05, 0.1\}.
\end{equation}
Conversely, to analyze the effect of graph size, we varied the number of nodes 
\begin{equation}
	N\in\{100 \cdot 2^i: i\in\{1, 2,3,4,5,6,7\} \},
\end{equation}
while keeping density fixed at $d=0.01$.
For each parameter configuration, we generated 10 graph instances per network model, aiming to reduce the influence of stochastic variation in graph generation. This setup allows us to analyze graph size and density separately, thereby reducing confounding effects between the two properties.

\subsection{Experimental Protocol}
Our experimental protocol consists of generating embeddings across multiple datasets and dimensions and subsequently assessing their representational stability. On the empirical datasets, we additionally assess functional stability and downstream predictive performance. In the following, we describe the corresponding procedures.

\subheader{Embedding Generation.}
To investigate the impact of dimensionality, we consider embedding dimensions
\begin{equation}
	D\in\{2^i, i\in \{2,3,\dots,12\}\}.
\end{equation}
For the empirical datasets, we generated 30 embeddings for each combination of embedding method, dataset, and dimensionality. For the synthetic datasets, we generated 10 graph instances for each combination of graph model and parameter configuration, and trained 10 embeddings per generated graph. That way, we account for variability arising from both stochastic graph generation and embedding training.

To ensure reasonable embedding quality without extensive over-optimization, we performed basic hyperparameter tuning \new{for every method--dataset combination at dimension $D=128$} prior to training the full range of embeddings across dimensions. \new{The selected hyperparameters were then reused across dimensions, keeping the training protocol fixed while varying dimensionality.}
\new{We briefly assess the sensitivity to this choice in Appendix~\ref{ap:hyperparameter_sensitivity}, where we perform dimension-specific tuning of embedding hyperparameters on the Wikipedia dataset.}
Additional details regarding hyperparameter tuning, implementation details, and parameter settings are provided in Appendix~\ref{ap:embedding_generation}. 

For a small number of configurations, embeddings could not be generated for practical reasons. Specifically, VERSE embeddings of dimension $D=4096$ on Cora consistently failed to compute despite repeated attempts. In addition, GraphSAGE embeddings on CoAuthor were only generated up to dimension $D=1024$, since training times exceeded ten hours per embedding at larger dimensions. 
For similar reasons, on the synthetic data, GraphSAGE embeddings were only computed for graphs with at most 3200 nodes.
Unless noted otherwise, all analyses were conducted using the successfully generated embeddings available for the respective configuration.

\subheader{Downstream Classification.}
We used the embeddings of the empirical datasets for downstream node classification and link prediction.
For both tasks, we applied a logistic regressor and a multi-layer perceptron (MLP). This was done to reduce the influence of classifier-specific effects and account for the additional stochasticity that neural downstream models may introduce during training.

For each dataset, we used fixed training, validation, and test splits, either based on existing benchmark protocols or sampled by us. Validation data was used for hyperparameter tuning of downstream classifiers, while test data was used to evaluate predictive performance and functional stability. Hyperparameter tuning was performed separately for each embedding dimension per dataset, to ensure a fair assessment of the representational capacity of lower-dimensional embeddings. For link prediction, node embeddings were preprocessed to classifier inputs using the Hadamard product, and negative edges were sampled uniformly from non-existing node pairs.

Further details regarding downstream classifiers, implementations, parameter grids, and downstream task construction are provided in Appendix \ref{ap:downstream_setup}.

\subsection{Evaluation}\label{sec:evaluation}
In our analyses, we need to evaluate the representational and functional stability of embedding models. We briefly present the measures that we use toward that end.

\subheader{Representational Stability.}
Following previous work~\citep{schumacher_effects_2021}, we selected \emph{aligned cosine similarity}~\citep{hamilton_diachronic_2016}, \emph{$k$-NN Jaccard similarity}~\cite{hellrich_bad_2016}, and \emph{second-order cosine similarity}~\citep{hamilton_cultural_2016} to evaluate representational stability. In addition, we applied \emph{distance correlation}~\citep{szekely_measuring_2007}, since this measure performed well consistently in the \resi{} benchmark for representational similarity measures~\citep{klabunde_resi_2025}. Collectively, these measures capture both global similarities between embedding spaces and the consistency of local neighborhood structure around individual node representations.

\emph{Aligned cosine similarity} captures global geometric similarity between embedding spaces by first aligning embeddings through the orthogonal Procrustes problem.
Specifically, the goal is to find the orthogonal matrix
\begin{equation}\label{eq:procrustes}
	\bm{Q^*} := \operatorname{argmin}_{\bm{Q} \in \operatorname{O}(D)} \|\Z \bm{Q} - \Zp\|_F,
\end{equation}
where $\operatorname{O}(D)$ denotes the orthogonal group of dimension $D$ and $\|\cdot\|_F$ the Frobenius norm. 
After aligning embeddings via $\bm{Q^*}$, one can then consider the cosine similarities between all node representations and use their average as similarity score:
\begin{equation}\label{eq:aligned_cossim}
	m_{\operatorname{AlignCos}}(\Z, \Zp) = \tfrac{1}{N}\textstyle\sum_{i=1}^N \operatorname{cos-sim}\left(\z_i \bm{Q}^*, \z'_i\right).
\end{equation}

Like aligned cosine similarity, distance correlation captures global differences between embedding spaces. However, it does so by comparing pairwise similarities between node representations rather than relying on explicit alignment.
These similarities are collected in a \emph{representational similarity matrix (RSM)} $\Sm\in \Real^{N \times N}$, which can be defined in terms of its elements via
\begin{equation}\label{eq:rsm}
	\Sm_{i,j}:=s(\z_i, \z_j),
\end{equation}
where  $s: \Real^D \times \Real^D \longrightarrow\Real$ denotes a given instance-wise similarity function.
Assuming that RSMs are mean-centered in both rows and columns, one can use their squared sample distance covariance $\operatorname{dCov}^2(\Sm,\Smp)=\tfrac{1}{N^2}\textstyle\sum_{i=1}^N\textstyle\sum_{j=1}^N \Sm_{i, j} \Sm'_{i,j}$ as similarity function $s$, and derive the distance correlation via 
\begin{equation}
	m_{\operatorname{DistCorr}}(\Z, \Zp)=\sqrt{\tfrac{\operatorname{dCov}^2(\Sm, \Smp)}{\sqrt{\operatorname{dCov}^2(\Sm, \Sm) \operatorname{dCov}^2(\Smp, \Smp)}}}.
\end{equation}

In contrast to the global measures above, \emph{$k$-NN Jaccard similarity} and \emph{second-order cosine similarity} focus on local neighborhood consistency in embedding space. 
Both measures determine the sets of $k$ nearest neighbors $\mathcal{N}^k_{\Z}(i)$ of each node's embedding $\z_i$ within the full embedding matrix $\Z$ with respect to cosine similarity and then compute vectors of node-wise neighborhood similarities $v_{\operatorname{NN}}(\Z,\Zp)\in\Real^N$, which are averaged over all nodes to obtain similarity measures for the full embeddings $\Z,\Zp$:
\begin{equation}
	m_{\operatorname{NN}}(\Z,\Zp) = \tfrac{1}{N} \textstyle\sum_{i=1}^N v_{\operatorname{NN}}(\Z,\Zp)_i.
\end{equation}
For the $k$-NN Jaccard similarity, this vector contains the Jaccard similarities of the nearest neighbors of each pair of corresponding node-level representations $\z_i$ and $\z'_i$:
\begin{equation} \label{eq:jaccard}
	\big(\bm{v}_\text{Jaccard}^k\big(\Z, \Zp\big)\big)_i :=
	\tfrac{|\mathcal{N}^k_{\Z}(i)\cap \mathcal{N}^k_{\Zp}(i)|}{|\mathcal{N}^k_{\Z}(i)\cup \mathcal{N}^k_{\Zp}(i)|}.    
\end{equation}

Second-order cosine similarity considers the union of the node-wise nearest neighbors in terms of cosine similarity as an ordered set $\{j_1,\dots, j_{K(i)}\}:=\mathcal{N}^k_{\Z}(i)\cup \mathcal{N}^k_{\Zp}(i)$, and then compares these cosine similarities to the nearest neighbors via
\begin{equation}\label{eq:2ndcos}
	\big(\bm{v}_\text{2nd-Cos}^k\big(\Z, \Zp\big)\big)_i:=  \quad \operatorname{cos-sim}\big(\big(\Sm_{i,j_1},\dots, \Sm_{i,j_{K(i)}}\big),\big(\Sm'_{i,j_1},\dots, \Sm'_{i,j_{K(i)}}\big)\big), \nonumber
\end{equation}
where $\Sm, \Smp$ denote the RSMs w.r.t. cosine similarity.

\subheader{Functional Stability.}
We evaluate functional stability primarily in terms of differences in individual downstream predictions, since prior work has shown that variability in aggregate performance measures such as accuracy or F1 score is typically low~\citep{schumacher_effects_2021}. To this end, we consider both measures based on hard class predictions and measures based on class-wise probability scores. Specifically, we evaluate \emph{disagreement}, \emph{min-max-normalized disagreement}, the \emph{stable core}, and \emph{Jensen-Shannon divergence}.

\emph{Disagreement}~\citep{madani_covalidation_2004} measures the rate of conflicting hard predictions between two outputs.
Letting $\Out, \Outp \in\Real^{n\times C}$ denote the output predictions resulting from two embeddings, $\yhat, \yhat'\in\Real^n$ the vectors of the corresponding hard predictions (as defined in Section \ref{sec:prelim}), and $\delta_{ij}$ the Kronecker delta, disagreement can be defined as
\begin{equation}\label{def:disagreement}
	m_{\operatorname{Dis}}(\Out, \Outp) = \tfrac{1}{N} \cdot \textstyle\sum_{i=1}^{N} \big(1 - \delta_{\yhat_i,\yhat_i'}\big) . 
\end{equation}

Given that disagreement is implicitly bounded by the error rates of classifiers, we further consider the \emph{min-max-normalized disagreement} \citep{klabunde_prediction_2022}, which, given the error rates $q_{\operatorname{Err}}(\cdot)$, relates the observed disagreement $m_{\operatorname{Dis}}(\Out, \Outp)$ to the respective minimum and maximum disagreement $m_{\operatorname{Dis}}^{(\min)}(\Out, \Outp) = |q_{\operatorname{Err}}(\Out) -  q_{\operatorname{Err}}(\Outp)|$ and $m_{\operatorname{Dis}}^{(\max)}(\Out, \Outp) = \min( q_{\operatorname{Err}}(\Out) + q_{\operatorname{Err}}(\Outp), 1)$.
This yields the measure
\begin{equation}
	m_{\operatorname{MinMaxNormDis}}(\Out, \Outp) = 
	\tfrac{m_{\operatorname{Dis}}(\Out, \Outp) - m_{\operatorname{Dis}}^{(\min)}(\Out, \Outp)}{m_{\operatorname{Dis}}^{(\max)}(\Out, \Outp) - m_{\operatorname{Dis}}^{(\min)}(\Out, \Outp)}.
\end{equation}

Disagreement only compares pairs of outputs, but high pairwise similarity does not necessarily imply consistency across a larger set of predictions $\mathcal{O}:=\{\Out, \Outp, \bm{O''}, \dots\}$.
Thus, we also consider the \emph{stable core}~\citep{schumacher_effects_2021}, which corresponds to the fraction of instance-wise predictions that agree across the whole group of outputs:
\begin{equation}\label{def:stable_core}
	m_{\text{StableCore}}(\mathcal{O}) =
	\tfrac{1}{N}\cdot\textstyle\sum_{i=1}^N \min_{\substack{\Out,\Outp \in \mathcal{O} \\ s.t.~\Out \neq \Outp}} \delta_{{\yhat}_i,{\yhat'}_i}. 
\end{equation}

In contrast to the previous measures, the \emph{Jensen-Shannon divergence} (JSD) evaluates differences in the full class-wise probability distributions averaged across instances.
Letting $\operatorname{KL}(\cdot \| \cdot)$ denote the Kullback-Leibler divergence, it is defined as
\begin{equation}
	m_{\text{JSD}}(\Out, \Outp) =  \tfrac{1}{2N}\textstyle\sum_{i=1}^N
	\operatorname{KL}(\Out_i \| \ols{\Out}_i) +  \operatorname{KL}(\Outp_i \| \ols{\Out}_i),
\end{equation}
where $\ols{\Out} = \frac{\Out + \Outp}{2}$ denotes the average output matrix.

\begin{revisionblock}
	\subheader{Bootstrap Analysis of Optimal-dimension Alignment.}
	To assess whether stability-optimal dimensions align with dimensions yielding near-optimal downstream performance, we generated 10,000 bootstrap replicates for each combination of dataset, embedding method, and stability measure.
	Each replicate comprised a separate bootstrap sample at every embedding dimension: at each dimension, the 30 runs were sampled with replacement, independently of the samples drawn at the other dimensions. 
	Within a dimension, the same resampled runs were used to estimate downstream performance and stability.
	
	Pairwise similarity, disagreement, or divergence scores were averaged only over pairs referring to \emph{different} original embeddings. 
	If embeddings $\Z$ and $\Z'$ occurred $c$ and $c'$ times in a bootstrap sample, respectively, their pairwise score $m(\Z,\Z')$ therefore contributed $c\cdot c'$ times to the average.
	Self-comparisons $m(\Z,\Z)$ induced by repeated selections of the same embedding were excluded, as their inclusion would artificially increase estimated stability.
	
	For each replicate and stability measure, we determined the stability-optimal dimension---maximizing similarity or minimizing disagreement or divergence---and the set of dimensions whose downstream accuracy was within
	$0.01$ of the replicate-specific maximum. 
	For each dataset, embedding method, and stability measure, we then calculated the \emph{bootstrap alignment rate} as the fraction of replicates in which the stability-optimal dimension belonged to this near-optimal-performance set.
\end{revisionblock}

\section{Results}\label{sec:results}

In the following, we present our results. We begin with downstream performance before analyzing the impact of dimensionality on representational and functional stability.
\new{Afterward, we analyze whether dimensions that yield the best performance align with the dimensions that the yield highest stability.}
Finally, we investigate how graph properties such as network size and density influence the relationship between embedding dimensionality and stability.

\begin{figure}[t]
	\centering
	\includegraphics[width=\textwidth]{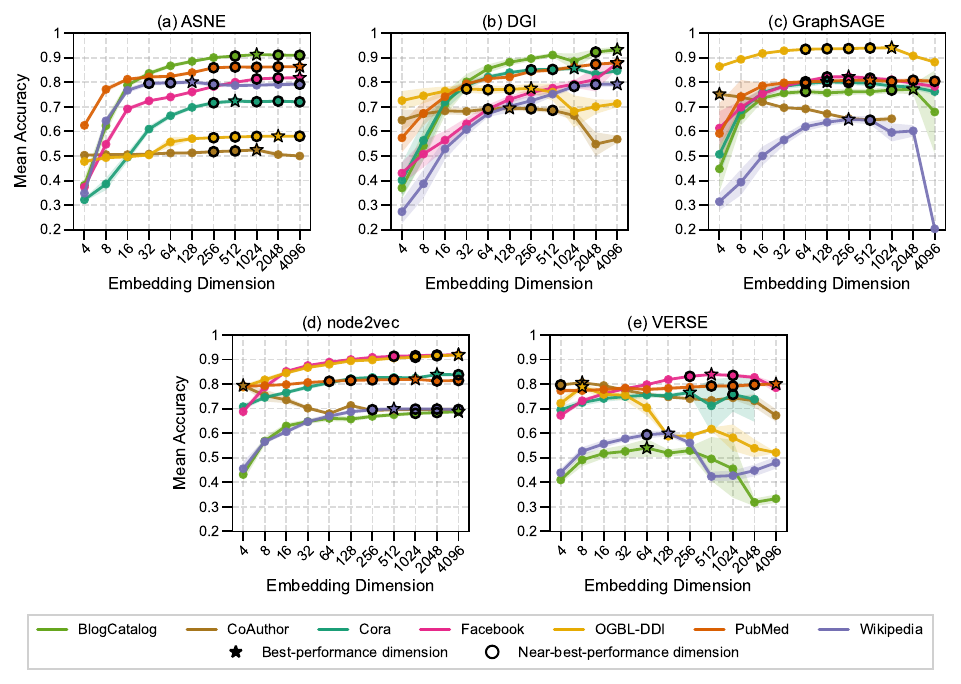}
	\caption{\captionheader{Impact of dimension on downstream performance.} 
		We present the average accuracy of each embedding method across 30 training runs using a logistic regressor as downstream classifier. \new{Shaded bands indicate one standard deviation around the respective means.} Stars denote the best-performing dimension for each dataset, dark rings indicate dimensions whose performance is within 0.01 of the optimum. Across most methods and datasets, performance improves with increasing dimensionality before reaching a plateau. Yet, for GraphSAGE and VERSE, very high dimensions often lead to deteriorating performance.
	}
	\label{fig:downstream_logreg}
\end{figure}

\subsection{Downstream Performance}\label{sec:performance}

We first analyze downstream performance across varying dimensions, which is presented in \Cref{fig:downstream_logreg}, where we use a logistic regressor as downstream classifier. Results obtained with an MLP are reported in \Cref{fig:downstream_mlp}; these exhibit largely similar trends.

Consistent with previous results \citep{schumacher_effects_2021, goyal_graph_2018, chen_graph_2020}, \new{variance in performance scores at fixed dimensions is generally very low, and} performance generally improves substantially when moving beyond very low dimensions.
Beyond this initial improvement, performance either plateaus or, for some embedding methods, deteriorates again at larger dimensions. ASNE, DGI, and node2vec most frequently follow the former pattern, often reaching broad performance plateaus with many dimensions yielding near-optimal results. GraphSAGE and VERSE, on the other hand, more often exhibit a pronounced peak followed by performance deterioration at higher dimensions.

The link prediction datasets show more heterogeneous trends than the node classification datasets. For DGI and particularly VERSE, optimal performance is often attained at comparatively small dimensions, after which it deteriorates. By contrast, node2vec and GraphSAGE do not follow a consistent task-specific pattern: while performance on OGBL-DDI resembles that observed on several node classification datasets, CoAuthor frequently favors substantially smaller dimensions.

ASNE behaves markedly differently on the link prediction datasets, where performance remains close to the prevalence of positive edges in the test data across all dimensions, corresponding to 50\% accuracy on CoAuthor and 59\% accuracy on OGBL-DDI. Inspection of the predictions revealed confidence scores close to 0.5 for most node pairs, with predictions appearing either nearly random or fully positive. This suggests that the learned embeddings provided little useful signal for link prediction, and we therefore exclude the corresponding results from our analysis of functional stability.

\newcommand{\jacASNE}{\trendglyph{(0,.08) (.18,.15) (.38,.38) (.58,.61) (.78,.69) (1,.68)}}
\newcommand{\jacDGI}{\heteroglyph{.05}{.31}{.60}}
\newcommand{\jacSAGE}{\trendglyph{(0,.02) (.18,.12) (.38,.23) (.60,.26) (.80,.22) (1,.17)}}
\newcommand{\jacNtwoV}{\trendglyph{(0,.23) (.16,.43) (.36,.65) (.60,.74) (.80,.76) (1,.76)}}
\newcommand{\jacVERSE}{\trendglyph{(0,.01) (.18,.06) (.40,.17) (.60,.22) (.76,.19) (1,.05)}}

\newcommand{\socASNE}{\trendglyph{(0,.08) (.16,.42) (.34,.78) (.52,.95) (.72,.99) (1,.99)}}
\newcommand{\socDGI}{\trendglyph{(0,.02) (.18,.18) (.38,.54) (.58,.79) (.74,.84) (.86,.75) (1,.88)}}
\newcommand{\socSAGE}{\trendglyph{(0,.05) (.16,.32) (.34,.61) (.50,.66) (.68,.62) (.84,.55) (1,.46)}}
\newcommand{\socNtwoV}{\trendglyph{(0,.28) (.14,.58) (.30,.82) (.48,.96) (.68,.99) (1,.99)}}
\newcommand{\socVERSE}{\trendglyph{(0,.04) (.15,.31) (.32,.70) (.50,.91) (.70,.98) (1,.99)}}

\newcommand{\cosASNE}{\trendglyph{(0,.84) (.16,.77) (.32,.66) (.44,.62) (.58,.69) (.74,.86) (.88,.95) (1,.96)}}
\newcommand{\cosDGI}{\trendglyph{(0,.97) (.35,.96) (.62,.90) (.78,.84) (.88,.86) (.96,.98) (1,.99)}}
\newcommand{\cosSAGE}{\partialglyph{(0,.40) (.15,.63) (.30,.68) (.48,.66) (.68,.63) (.84,.59) (1,.56)}}
\newcommand{\cosNtwoV}{\trendglyph{(0,.96) (.20,.94) (.42,.90) (.60,.93) (.80,.98) (1,.98)}}
\newcommand{\cosVERSE}{\trendglyph{(0,.83) (.20,.66) (.43,.28) (.60,.29) (.82,.39) (1,.43)}}

\newcommand{\distASNE}{\trendglyph{(0,.87) (.18,.83) (.36,.74) (.50,.76) (.68,.88) (.84,.96) (1,.97)}}
\newcommand{\distDGI}{\heteroglyph{.24}{.55}{.91}}
\newcommand{\distSAGE}{\trendglyph{(0,.48) (.15,.72) (.28,.81) (.48,.79) (.68,.75) (.84,.70) (1,.65)}}
\newcommand{\distNtwoV}{\trendglyph{(0,.90) (.18,.94) (.38,.97) (.58,.99) (.78,.995) (1,.995)}}
\newcommand{\distVERSE}{\trendglyph{(0,.78) (.16,.61) (.34,.45) (.52,.53) (.68,.66) (.84,.34) (1,.68)}}

\begin{table}[t]
\centering

\begingroup
\footnotesize

\setlength{\tabcolsep}{3pt}
\renewcommand{\arraystretch}{1.30}

\renewcommand{\tabularxcolumn}[1]{m{#1}}

\begin{tabularx}{\linewidth}{
  @{}
  L{12mm}
  *{4}{Y}
  @{}
}
\toprule
&
\multicolumn{2}{c}{\textbf{Local Stability}} &
\multicolumn{2}{c}{\textbf{Global Stability}}
\\

\cmidrule(lr){2-3}
\cmidrule(lr){4-5}

&
\shortstack[c]{%
  \textbf{$k$-NN Jaccard}\\
  \textbf{Similarity}
}
&
\shortstack[c]{%
  \textbf{Second-order}\\
  \textbf{Cosine Similarity}
}
&
\shortstack[c]{%
  \textbf{Aligned Cosine}\\
  \textbf{Similarity}
}
&
\shortstack[c]{%
  \textbf{Distance}\\
  \textbf{Correlation}
}
\\

\midrule
\methodlabel{ASNE} &
  \cell{\jacASNE}{6/7} &
  \cell{\socASNE}{7/7} &
  \cell{\cosASNE}{7/7} &
  \cell{\distASNE}{5/7} \\
\addlinespace[1.2mm]
\methodlabel{DGI} &
  \heterocell{\jacDGI} &
  \cell{\socDGI}{5/7} &
  \cell{\cosDGI}{5/7} &
  \heterocell{\distDGI} \\
\methodlabel{\shortstack[l]{Graph-\\SAGE}} &
  \cell{\jacSAGE}{7/7} &
  \cell{\socSAGE}{7/7} &
  \partialcell{\cosSAGE}{4/7} &
  \cell{\distSAGE}{5/7} \\
\addlinespace[1.2mm]
\methodlabel{node2vec} &
  \cell{\jacNtwoV}{5/7} &
  \cell{\socNtwoV}{7/7} &
  \cell{\cosNtwoV}{5/7} &
  \cell{\distNtwoV}{5/7} \\
\addlinespace[1.2mm]
\methodlabel{VERSE} &
  \cell{\jacVERSE}{5/7} &
  \cell{\socVERSE}{7/7} &
  \cell{\cosVERSE}{7/7} &
  \cell{\distVERSE}{7/7} \\
\bottomrule
\end{tabularx}
\endgroup
\smallskip
\caption{\new{
\captionheader{Impact of dimensionality  on representational stability.}
For each embedding method (rows) and similarity measure (columns), we present schematic curves summarizing how representational stability changes with dimensionality.
In each plot, the x-axis represents dimension (increasing exponentially from 4 to 4096), and the y-axis represents similarity scores. 
All four measures range from $0$ to $1$, with higher values indicating greater stability.
The solid gray lines mark $0.5$. 
The curves show both trend shape and approximate similarity levels.
Solid teal curves (\legendtealline{}) indicate consensus trends observed for at least two-thirds of the available datasets, while dashed teal curves (\legendtealdashedline{}) indicate partial trends observed for more than half of the observed datasets. Crossing gray curves (\legendgrayx{}) indicate that no single recurring trend emerged, but their vertical extent nevertheless reflects the approximate range of observed similarity scores. Fractions report the number of datasets supporting the depicted trend relative to the number available for the corresponding method--measure combination. Overall trends are fairly diverse, with local stability often increasing with higher dimensions, and global stability largely displaying non-monotonic trends.
}
}\label{tab:summary_repsim}

\end{table}

\subsection{Representational Stability}\label{sec:repsim}

We present a summary of the recurring dimensionality-dependent trends for representational stability in \Cref{tab:summary_repsim}, while the detailed plots which include dataset-specific curves are provided in Appendix~\ref{ap:representational}.

For local stability, we observe that dimensionality-dependent trends overall tend to align across both measures, with VERSE being the only clear exception to this pattern: its $k$-NN Jaccard similarity remains comparatively low and tends to peak at intermediate dimensions, whereas second-order cosine similarity increases toward a high plateau.
ASNE and node2vec generally become more stable as dimensionality increases, reaching comparatively high plateaus under both measures.
DGI also tends to become more stable toward higher dimensions under both measures. For $k$-NN Jaccard similarity, however, this ranges from pronounced increases toward moderate stability levels to only weak changes at low levels (cf. \Cref{fig:jaccard}).
GraphSAGE shows a different but consistent pattern across both measures: stability increases toward a comparatively low peak at intermediate dimensions before declining again.

However, importantly, the stated agreement in trend shape does not necessarily imply agreement in the indicated stability level: except for GraphSAGE, second-order cosine similarity generally tends to saturate near its maximum possible value of one, while $k$-NN Jaccard similarity hardly ever gets close to its maximum.
Such differences between local measures have previously been reported for fixed-dimensional node2vec and GraphSAGE embeddings~\citep{schumacher_effects_2021}. Our results additionally show that their magnitude can vary across embedding dimensions.

Global stability exhibits predominantly non-monotonic relationships with dimensionality.
node2vec generally remains highly stable across dimensions and often approaches near-perfect similarity at larger dimensions.  
ASNE instead commonly decreases in its stability down to a minimum at an intermediate dimension before recovering toward high stability. 
For DGI, aligned cosine similarity generally remains relatively high but often dips at larger dimensions, whereas distance correlation exhibits no recurring trend.
GraphSAGE typically increases in stability initially, reaching a peak at a moderate to high stability level before gradually declining at larger dimensions. This pattern is, however, less consistent for aligned cosine similarity. 
For VERSE, similarity scores consistently decrease from initially high values at low dimensionality before partially recovering at higher dimensions. With respect to distance correlation, this recovery is often interrupted by additional dips in stability.

Considering similarities to the downstream performance curves, we find that in particular for local stability measures, these are often resembled via the increases toward plateaus for ASNE, DGI, and node2vec, and peaks at intermediate dimensions for GraphSAGE. 
This correspondence is, however, not uniform across measures, as illustrated most clearly by VERSE, where only $k$-NN Jaccard similarity yields similar trends as downstream performance.
The correspondence with downstream performance is less apparent for the global stability measures, whose non-monotonic trajectories frequently continue changing after performance has plateaued.
We note that these comparisons concern broad curve shapes and do not imply that the corresponding optimal dimensions coincide---their alignment is examined separately in \Cref{sec:dimension_alignment}.

\newcommand{\coreASNE}{\trendglyph{(0,.15) (.20,.32) (.42,.57) (.62,.72) (.82,.82) (1,.82)}}
\newcommand{\coreDGI}{\trendglyph{(0,.02) (.20,.13) (.42,.39) (.62,.61) (.77,.65) (.88,.59) (1,.70)}}
\newcommand{\coreSAGE}{\trendglyph{(0,.05) (.18,.28) (.35,.43) (.58,.45) (.75,.41) (1,.27)}}
\newcommand{\coreNtwoV}{\trendglyph{(0,.55) (.20,.61) (.42,.71) (.62,.80) (.82,.85) (1,.85)}}
\newcommand{\coreVERSE}{\partialglyph{(0,.20) (.16,.27) (.34,.37) (.52,.45)(.64,.46) (.76,.38) (.88,.25) (1,.12)}}

\newcommand{\disASNE}{\trendglyph{(0,.52) (.18,.40) (.38,.26) (.58,.15) (.78,.08) (1,.07)}}
\newcommand{\disDGI}{\trendglyph{(0,.55) (.18,.43) (.38,.27) (.58,.17) (.78,.12) (1,.10)}}
\newcommand{\disSAGE}{\trendglyph{(0,.48) (.16,.31) (.34,.20) (.58,.18) (.80,.19) (1,.24)}}
\newcommand{\disNtwoV}{\trendglyph{(0,.36) (.18,.27) (.38,.19) (.58,.11) (.78,.07) (1,.06)}}
\newcommand{\disVERSE}{\trendglyph{(0,.38) (.25,.25) (.4,.18) (.6,.24) (.8,.34) (1,.44)}}

\newcommand{\nmdASNE}{\partialglyph{(0,.14) (.15,.28) (.32,.43) (.52,.38) (.72,.28) (1,.18)}}
\newcommand{\nmdDGI}{\heteroglyph{.27}{.45}{.64}}
\newcommand{\nmdSAGE}{\heteroglyph{.11}{.35}{.59}}
\newcommand{\nmdNtwoV}{\heteroglyph{.05}{.22}{.39}}
\newcommand{\nmdVERSE}{\heteroglyph{.18}{.47}{.75}}

\newcommand{\jsdASNE}{\heterologglyph{.038}{.218}{.413}}
\newcommand{\jsdDGI}{\heterologglyph{.173}{.338}{.510}}
\newcommand{\jsdSAGE}{\heterologglyph{.225}{.375}{.525}}
\newcommand{\jsdNtwoV}{\heterologglyph{.015}{.203}{.413}}
\newcommand{\jsdVERSE}{\heterologglyph{.015}{.338}{.675}}

\begin{table}[t]
	\centering
	
	\begingroup
	\footnotesize
	
	\setlength{\tabcolsep}{3pt}
	\renewcommand{\arraystretch}{1.30}
	
	\renewcommand{\tabularxcolumn}[1]{m{#1}}
	
	\begin{tabularx}{\linewidth}{
			@{}
			L{12mm}
			*{4}{Y}
			@{}
		}
		\toprule
		&
		\multicolumn{3}{c}{\raisebox{1.2ex}{\textbf{Hard Predictions}}} &
		\multicolumn{1}{c}{\shortstack{\textbf{Predicted} \\ \textbf{Probabilities}}} \\
		\cmidrule(lr){2-4}\cmidrule(lr){5-5}
		&
		\shortstack{\textbf{Stable core}\\\textbf{\((\uparrow)\)}} &
		\shortstack{\textbf{Disagreement}\\\textbf{\((\downarrow)\)}} &
		\shortstack{\textbf{Min-Max-Norm.}\\\textbf{Disagreement \((\downarrow)\)}} &
		\shortstack{\textbf{Jensen-Shannon} \\ \textbf{Divergence} \((\downarrow)\)} \\
		\midrule
		\methodlabel{ASNE} &
		\cell{\coreASNE}{4/5} &
		\cell{\disASNE}{5/5} &
		\partialcell{\nmdASNE}{3/5} &
		\heterocell{\jsdASNE} \\
		\addlinespace[1.2mm]
		\methodlabel{DGI} &
		\cell{\coreDGI}{5/7} &
		\cell{\disDGI}{5/7} &
		\heterocell{\nmdDGI} &
		\heterocell{\jsdDGI} \\
		\methodlabel{\shortstack[l]{Graph-\\SAGE}} &
		\cell{\coreSAGE}{5/7} &
		\cell{\disSAGE}{5/7} &
		\heterocell{\nmdSAGE} &
		\heterocell{\jsdSAGE} \\
		\addlinespace[1.2mm]
		\methodlabel{node2vec} &
		\cell{\coreNtwoV}{7/7} &
		\cell{\disNtwoV}{6/7} &
		\heterocell{\nmdNtwoV} &
		\heterocell{\jsdNtwoV} \\
		\addlinespace[1.2mm]
		\methodlabel{VERSE} &
		\partialcell{\coreVERSE}{4/7} &
		\cell{\disVERSE}{6/7} &
		\heterocell{\nmdVERSE} &
		\heterocell{\jsdVERSE} \\
		\bottomrule
	\end{tabularx}
	\endgroup
	\smallskip
	\caption{\new{
			\captionheader{Impact of dimensionality on functional stability.}
			For each embedding method (rows) and stability measure (columns), we present schematic curves summarizing how functional stability changes with dimensionality.
			In each plot, the x-axis represents dimension (increasing exponentially from 4 to 4096), and the y-axis represents similarity or divergence scores. 
			Arrows in the column headers indicate whether higher (\(\uparrow\)) or lower (\(\downarrow\)) values correspond to greater stability.
			All measures range from 0 to 1, and the solid gray lines mark 0.5.
			JSD uses a logarithmic scale, where dashed gray lines mark successive powers of ten, down to $10^{-4}$.
			The curves show both trend shape and approximate values of the corresponding measure.
			Solid teal curves (\legendtealline{}) indicate consensus trends observed for at least two-thirds of the available datasets, while dashed teal curves (\legendtealdashedline{}) indicate partial trends observed for more than half but fewer than two-thirds of them.
			Crossing gray curves (\legendgrayx{}) indicate that no single recurring trend emerged, but their vertical extent nevertheless reflects the approximate range of observed values.
			Fractions report the number of datasets supporting the depicted trend relative to the number available for the corresponding method--measure combination.
			Stable core and disagreement show corresponding results, with stability often increasing across dimensions. The other measures hardly show consistent trends. 
		}
	}\label{tab:summary_funcsim}
\end{table}

\subsection{Functional Stability}\label{sec:funcsim}

We present the recurring dimensionality-dependent trends and approximate levels of functional stability in \Cref{tab:summary_funcsim}. 
The complete results obtained with logistic regression are provided in Appendix~\ref{ap:funcsim_logreg}. Results obtained with an MLP are reported in Appendix~\ref{ap:funcsim_mlp} and exhibit largely similar trends, although instability at very high dimensions is occasionally more pronounced.

It can be seen that stable core and disagreement yield broadly corresponding method-level trends, as may be expected intuitively.
For ASNE, DGI, and node2vec, stable core values generally increase with dimensionality toward comparatively high plateaus, while disagreement decreases toward low levels. GraphSAGE also becomes more stable initially, but both measures indicate a slight loss of stability at larger dimensions. 
For VERSE, stable core frequently increases before declining at high dimensions, although this trend is only partially supported across datasets. Disagreement more consistently exhibits the inverse trajectory, frequently reaching an intermediate-dimensional minimum before increasing again. The magnitude of this increase, however, varies considerably across datasets (cf. \Cref{fig:disagreement}).

Min-max-normalized disagreement does not reproduce the recurring trends observed for unnormalized disagreement, since apart from a partial trend for ASNE, its relationship with dimensionality varies across datasets for all embedding methods. 
Further, Jensen-Shannon divergence also reveals no recurring dimensionality-dependent trend while remaining at very low levels on its logarithmic scale. 

Broad similarities with the corresponding performance curves are therefore most apparent for the stable core and unnormalized disagreement. For these measures, initial performance gains are often accompanied by increasing stability, while performance declines at high dimensions frequently coincide with reduced stability.
Whether these similarities in broad curve shape also result in alignment between optimal dimensions is examined in \Cref{sec:dimension_alignment}.

\begin{figure}[t!]
	\centering
	\includegraphics[width=\textwidth]{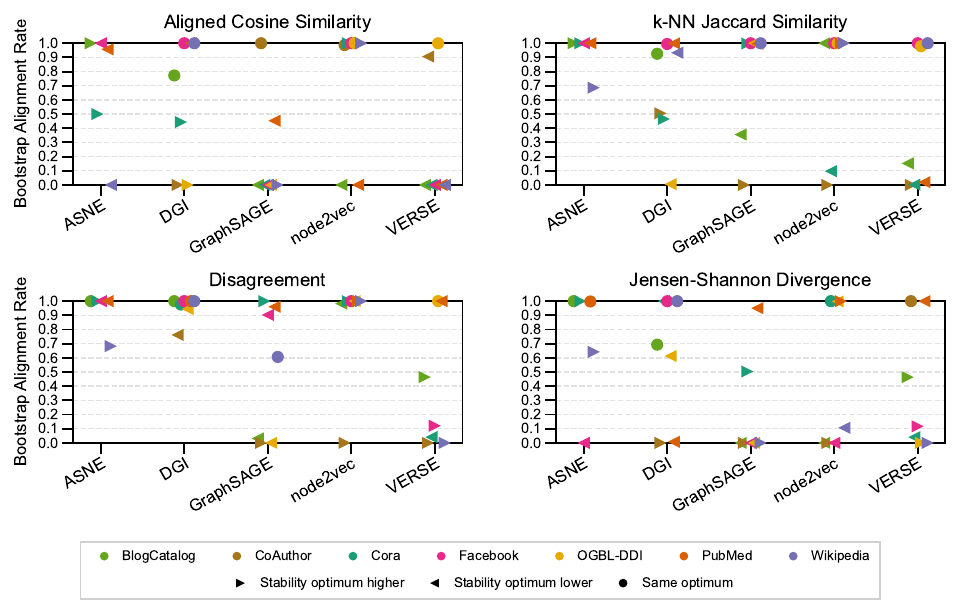}
	\caption{\new{\captionheader{Bootstrap alignment of stability-optimal dimensions with dimensions yielding near-optimal performance.}
			Each point shows, for one embedding method, stability measure, and dataset, the fraction of 10,000 bootstrap replicates in which the stability-optimal dimension is among the dimensions yielding downstream accuracy within 0.01 of the replicate-specific maximum.
			Colors identify datasets.
			Right- and left-pointing triangles indicate whether, using all runs, the stability optimum occurs at a higher or lower dimension than the performance optimum, respectively. Circles indicate coinciding optima.
			Dimensions yielding the highest performance do not necessarily coincide with those yielding most stable embeddings, though results vary strongly across datasets, embedding methods and similarity measures. 
	}}
	\label{fig:bootstrap_alignment}
\end{figure}

\subsection{Alignment Between Optimal Dimensions for Stability and Performance}\label{sec:dimension_alignment}

We assessed the alignment of stability- and performance-optimal dimensions using the bootstrap procedure described in \Cref{sec:evaluation}. To cover each of the four stability notions distinguished in our evaluation, \Cref{fig:bootstrap_alignment} presents the bootstrap alignment rates for aligned cosine similarity as a global representational measure, $k$-NN Jaccard similarity as a local representational measure, disagreement between hard predictions, and JSD between predictive probability distributions. Results for the remaining measures are reported in Appendix~\ref{ap:bootstrap}.

We observe that most bootstrap alignment rates are close to zero or one, showing that the bootstrap replicates usually agree on whether the stability-optimal dimension is also near-optimal in performance. Across configurations, however, alignment varies substantially, and cases of consistently high or low alignment across all datasets are rare. Among the shown measures, disagreement exhibits the overall highest alignment with near-optimal performance, although GraphSAGE and VERSE, which appear comparatively less aligned across the observed measures, also show mixed alignment for disagreement. This comparatively high alignment does not extend to min-max-normalized disagreement, as shown in \Cref{fig:bootstrap_alignment_ap}. For the remaining stability measures, no similarly clear alignment pattern emerges across datasets and embedding methods.

Further, even when alignment with near-optimal performance is high, the exact performance- and stability-optimal dimensions frequently differ. The strict-alignment results shown in \Cref{fig:bootstrap_alignment_strict} additionally demonstrate that these exact mismatches generally persist across bootstrap replicates. The contrast between near-optimal and strict alignment is consistent with the broad performance plateaus visible in \Cref{fig:downstream_logreg}: the detailed curves of stability across dimensions in Appendices~\ref{ap:representational} and \ref{ap:funcsim_logreg} show that stability can continue changing across dimensions that yield nearly identical performance. However, no consistent tendency emerges for the stability optimum to occur at a lower or higher dimension than the performance optimum.

Overall, we conclude that stability- and performance-optimal dimensions do not necessarily coincide, but neither is their divergence universal.

\subsection{Impact of Graph Properties}
Finally, we utilize the synthetic datasets to analyze how graph size and density influence the relationship between embedding dimensionality and stability.

\begin{figure}[t!]
	\centering
	\includegraphics[width=\textwidth]{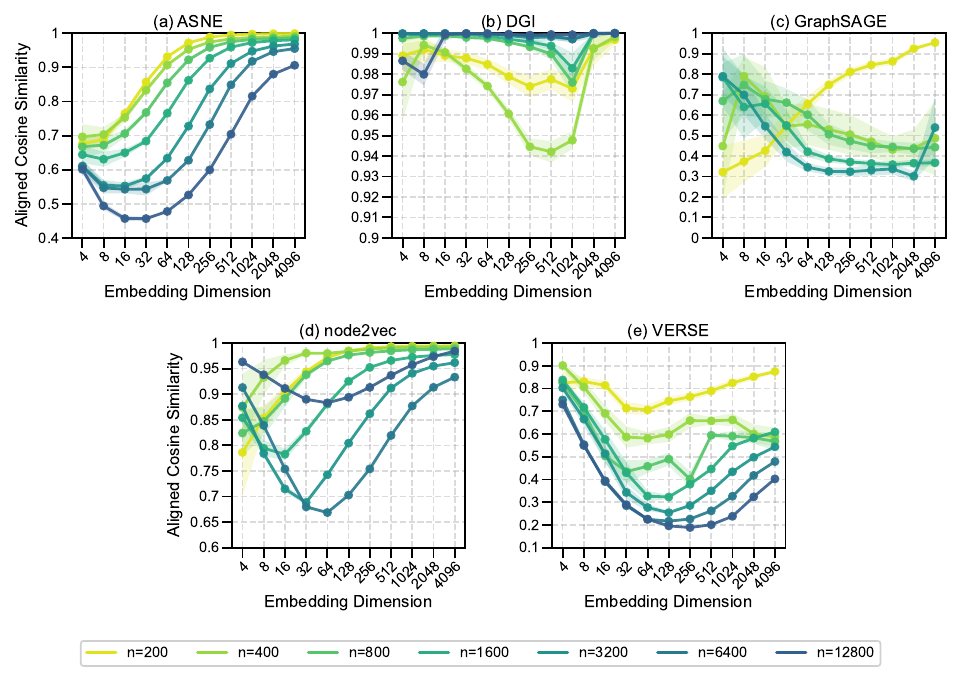}
	\caption{\captionheader{Interplay between network size and embedding dimension on Barabasi-Albert graphs.}
		We depict aligned cosine similarity averaged over all pairs of 10 embeddings of same graphs, taken across 10 different instantiations of the Barabasi-Albert graph at density $d=0.01$, with varying size as indicated by line color. \new{Shaded bands indicate one standard deviation around the respective means.} 
		Higher values indicate more stable embeddings.
		For ASNE, node2vec, and VERSE, larger networks generally require higher embedding dimensions before stable high-dimensional regimes are reached. By contrast, trends for DGI and GraphSAGE are less consistent.
	}
	\label{fig:ba_size_aligned_cossim}
\end{figure}

\subheader{Impact of Network Size.}
\Cref{fig:ba_size_aligned_cossim} depicts representational stability in terms of aligned cosine similarity for Barabási-Albert graphs of varying size and fixed density across different embedding dimensions. Corresponding results for additional similarity measures and Watts-Strogatz graphs are provided in Appendix~\ref{ap:size}.

Overall, the observed dimensionality-dependent trends broadly resemble those discussed for empirical datasets in \Cref{sec:repsim}. In particular, ASNE, node2vec, and VERSE frequently exhibit non-monotonic stability curves in which stability initially decreases before recovering again at higher dimensions. Across these methods, increasing network size generally shifts both the stability minima and the transition toward stable regimes to larger dimensions. In addition, smaller graphs often exhibit higher stability overall across most dimensions. These patterns are particularly consistent for node2vec and remain broadly visible across both graph models and multiple similarity measures (cf. Appendix~\ref{ap:size}). For VERSE, however, the recovered high-dimensional stability levels generally remain substantially lower than for ASNE and node2vec.

By contrast, results for DGI and GraphSAGE are less consistent. DGI nevertheless often exhibits comparatively high stability across dimensions, although occasional sharp decreases at intermediate or high dimensions occur depending on the graph model and similarity measure. GraphSAGE also does not display a clear size-dependent trend beyond the observation that the smallest graphs ($N=200$) often become increasingly stable with growing dimensionality.

\begin{figure}[t]
	\centering
	\includegraphics[width=\textwidth]{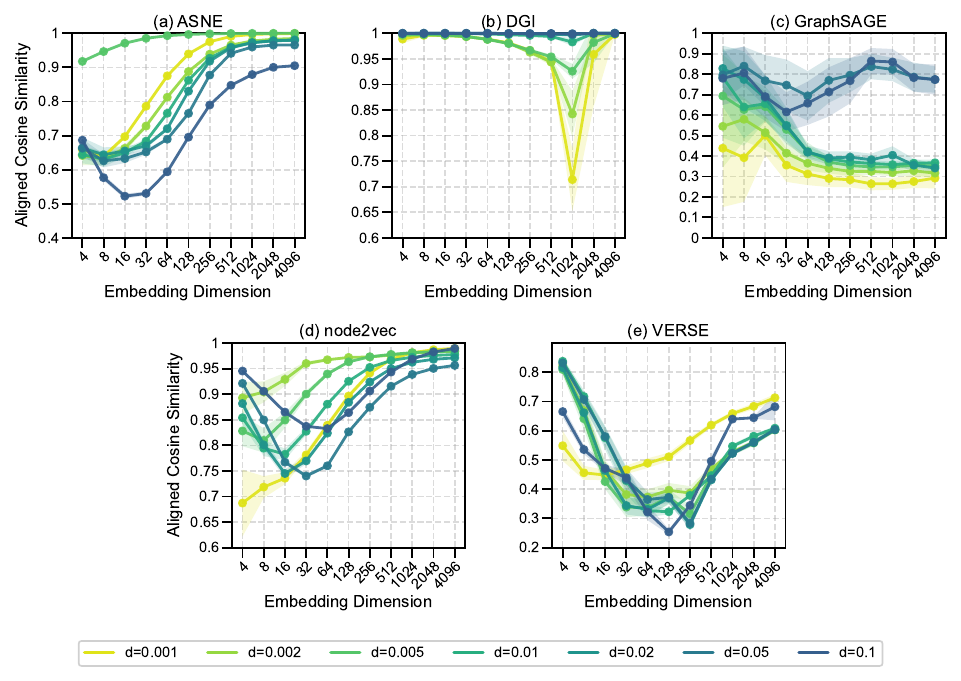}
	\caption{\captionheader{Interplay between network density and embedding dimension on Barabasi-Albert graphs.}
		We depict aligned cosine similarity averaged across all pairs of 10 embeddings computed on the same graph, aggregated over 10 independently generated Barabasi-Albert graphs with $N=1600$ nodes and varying density as indicated by line color. \new{Shaded bands indicate one standard deviation around the respective means.} Higher values indicate more stable embeddings. While some density-dependent patterns appear for individual methods, these trends are not as consistent across graph models and similarity measures as those observed for graph size.
	}
	\label{fig:ba_density_aligned_cossim}
\end{figure}

\subheader{Impact of Density.}
\Cref{fig:ba_density_aligned_cossim} depicts representational stability in terms of aligned cosine similarity for Barabasi-Albert graphs of varying density and fixed size across embedding dimensions. Corresponding results for additional similarity measures and Watts-Strogatz graphs are provided in Appendix~\ref{ap:density}.

Overall, the dimensionality-dependent stability curves again broadly resemble those observed for empirical datasets and in the graph-size analysis, particularly for ASNE, node2vec, and VERSE. DGI again exhibits pronounced downward spikes in stability around dimension 1024.

In contrast to graph size, the interplay between density, dimensionality, and stability appears substantially less consistent across methods, graph models, and similarity measures. Some tendencies nevertheless recur more frequently across graph types and similarity measures. In particular, denser ASNE embeddings, and to a lesser extent node2vec and VERSE embeddings, sometimes require larger dimensions before stable high-dimensional regimes are reached. However, even for ASNE, the separation between sparse and dense graphs becomes much weaker on Watts-Strogatz networks and for several alternative similarity measures (cf. Appendix~\ref{ap:density}).

Most other observations that could be drawn from \Cref{fig:ba_density_aligned_cossim} do not generalize across graph models and similarity measures. For example, the stronger downward spikes of DGI embeddings for sparse graphs are not consistently reproduced on Watts-Strogatz networks. Similarly, denser GraphSAGE embeddings are not generally more stable than sparse embeddings. Overall, no relationship as systematic as that observed for graph size emerges.

\section{Discussion}\label{sec:dimpact_discussion}
In the following, we discuss our findings with respect to the research objectives, relate them to prior work, and highlight their implications and limitations.

\subheader{Summary of Main Findings.}
Our results show that embedding dimensionality can substantially affect both representational and functional stability. However, no universal relationship between dimensionality and stability emerged, as the observed effects differed across embedding methods and stability notions. Several broader patterns nevertheless recurred. Neighborhood stability and hard-prediction stability often exhibited similar responses to increasing dimensionality, suggesting that consistent local relationships may support consistent downstream predictions. Global representational stability predominantly exhibited non-monotonic relationships with dimensionality. For stability in predicted probabilities, we observed that JSD remained very low and showed no recurring dimensionality-dependent trends, even where the stability of hard predictions changed with dimensionality. This suggests that low probability divergence does not necessarily imply stable class predictions, since small run-to-run differences can alter predictions near the decision boundary and the frequency of such cases may vary with dimensionality.

Regarding the relationship between stability and downstream performance, stability- and performance-optimal dimensions did not necessarily align, although this divergence was not universal. For disagreement, stability-optimal dimensions frequently fell within the set of dimensions yielding near-optimal performance, which is partly expected because high predictive accuracy constrains disagreement. 
Even where alignment with near-optimal performance was high, the exact optima frequently differed, and these mismatches generally persisted across bootstrap replicates. This contrast is consistent with the frequently observed broad performance plateaus, along which stability may continue changing while performance remains nearly constant. Further, there was no consistent tendency as to whether stability optima occurred at lower or higher dimensions than performance optima. 
Correspondence with performance was also less apparent for global than for local representational stability, suggesting that changes in global embedding geometry need not affect task-relevant information.

Finally, regarding how graph properties influence dimensionality-dependent stability, the synthetic experiments indicate that increasing graph size generally shifted stable high-dimensional regimes toward larger dimensions. By contrast, no comparably consistent effect of density emerged across graph models and stability measures.

\subheader{Relation to Existing Results.}
Our findings broadly align with prior work showing that increasing embedding dimensionality typically improves downstream performance up to a threshold, after which performance either plateaus or deteriorates~\citep{goyal_graph_2018,chen_graph_2020}. We confirm this general picture but extend it beyond a performance-centric perspective by tracking representational and functional stability across dimensions.
That way, we also reveal that stability and performance do not necessarily evolve in tandem.

Prior work also suggested that embedding dimensionality influences stability. For example, \citet{wang_understanding_2022} identified dimensionality as a significant predictor of embedding stability in their regression analysis. Our results broadly reinforce this observation, while suggesting that the relationship between dimensionality and stability is often non-linear. Depending on the embedding method and stability notion considered, increasing dimensionality may lead to gradual stabilization, non-monotonic behavior, or delayed recovery of stability at larger dimensions.

Similarly, \citet{gu_principled_2021} observed that representational stability of node2vec embeddings initially decreases before recovering at higher dimensions. We reproduced this result and placed it into a broader context by showing that dimensionality-stability relationships can take diverse forms across embedding methods and stability notions. 

Finally, our findings also relate to recent work on intrinsic graph dimensionality and dimensionality selection methods such as MinGE~\citep{luo_graph_2021}, which aim to estimate compact embedding dimensions from graph structure. Interestingly, previously proposed intrinsic-dimensionality estimates and recommended embedding dimensions for several benchmark datasets often fall substantially below the dimensionality regimes in which we observed maximal stability or downstream performance (cf. Appendix~\ref{ap:min_ge}). This suggests that entropy-oriented dimensionality estimates capture properties that differ from those reflected in stability and downstream performance.

\subheader{Implications.}
Our results suggest that embedding dimensionality can affect downstream performance, representational stability, and predictive stability in substantially different ways. While these properties often improved together---particularly for methods such as node2vec, ASNE, and DGI---their optima did not necessarily coincide. For these methods, stability often continued improving beyond dimensionality regimes in which downstream performance had already plateaued, potentially incentivizing the use of larger dimensions.
At the same time, these apparent benefits would need to be weighed against the increased computational overhead associated with high-dimensional embeddings.
Importantly, these trends also did not generalize across all embedding methods and datasets, particularly for GraphSAGE and VERSE.
Consequently, no single dimensionality regime emerged as universally preferable, and the most suitable embedding dimension may depend not only on predictive performance, but also on the relative importance of stability and computational efficiency.

The synthetic experiments further suggest that dimensionality requirements may increase with network size, as larger graphs often reached stable high-dimensional regimes only at larger embedding dimensions. In practice, this indicates that dimensionality choices should be considered jointly with both the intended downstream objective and characteristics of the underlying graph.

\subheader{Limitations.}
The stability of the investigated methods may be influenced by their specific implementations.
For VERSE, we used the reference implementation by \citet{tsitsulin_verse_2018}, whereas the remaining methods were implemented using frameworks such as \texttt{GRAPE}~\citep{cappelletti_grape_2023}, \texttt{karateclub}~\citep{rozemberczki_karate_2020}, and \texttt{PyTorch Geometric}~\citep{fey_pyg_2025}, which we used for DGI and GraphSAGE. \new{Since we evaluated one implementation per method, we cannot disentangle method-specific from implementation-specific effects on stability.} 
Consequently, differences between our GraphSAGE results and prior stability studies~\citep{schumacher_effects_2021} may partially reflect differences in implementation and training setup, and similar effects may also occur for the other methods\new{, particularly ASNE, whose evaluated implementation differs from the original formulation}.
\new{More broadly, our experiments also do not determine which algorithmic mechanisms lead to changes in stability across dimensions or isolate the contributions of individual sources of randomness.}

\new{The main experiments reused embedding hyperparameters that were determined at dimension $D=128$. 
We performed a brief sensitivity analysis on Wikipedia (Appendix~\ref{ap:hyperparameter_sensitivity}), where tuning did not appear to substantially alter the general dimensionality-dependent trends.
Nevertheless, we cannot rule out that instability at extreme dimensions partly reflects parameter mismatch rather than the training dynamics themselves.}

Due to the computational cost of repeated embedding training \new{(cf. Appendix \ref{ap:costs})}, our experiments did not extend to very large graphs. In particular, higher-dimensional embeddings incur substantial computational overhead, constraining both the maximal dimensions and the granularity of dimensions we could evaluate. As a result, certain trends---especially in very high-dimensional regimes---may not have been fully captured.

Finally, the assessment of representational stability depends on the chosen similarity measures. While we considered a broad set of measures informed by prior work~\citep{schumacher_effects_2021, wang_dimensionality_2023, klabunde_resi_2025}, it remains possible that the selected measures did not always perfectly represent the geometry of embeddings produced by a specific method.
However, this concern does not apply to functional stability, where we evaluated predictive consistency across multiple granularity levels while also considering measures that control for agreement induced by high accuracies.

\section{Conclusion}
In this work, we conducted a large-scale analysis of how embedding dimensionality affects the stability of node embeddings. Across five embedding methods, multiple empirical and synthetic datasets, and a broad range of representational and functional similarity measures, we found that dimensionality can substantially influence embedding stability, often in ways that differ from its effect on downstream predictive performance. At the same time, the observed relationships proved highly dependent on both the embedding method and the stability notion considered.

\new{Future work could develop method-specific theoretical or mechanistic explanations for the observed dimensionality-dependent stability patterns.}
In addition, our empirical approach could be extended to analyze the stability of embeddings and its relationship with dimensionality on other kinds of network data, such as temporal, \new{signed}, heterogeneous, or knowledge graphs, for which separate embedding methods exist.
Similarly, one could extend this study to end-to-end graph learning models such as supervised graph neural networks and graph transformers.

\section*{Acknowledgements}
This work is supported by the Deutsche Forschungsgemeinschaft (DFG, German Research Foundation) under Grant No.~453349072.
The authors acknowledge support by the state of Baden-Württemberg through bwHPC and DFG through grant INST 35/1597-1 FUGG.

\FloatBarrier

\printbibliography

\appendix

\section{Additional Experimental Details}\label{ap:dimpact_params}

\subsection{Datasets}\label{ap:datasets}
As described in the main text, we consider the citation networks \textit{Cora} and \textit{PubMed}~\citep{yang_revisiting_2016a}; the social networks \textit{BlogCatalog}~\citep{yang_pane_2023} and \textit{Facebook}~\citep{rozemberczki_multiscale_2021}; the hyperlink network \textit{Wikipedia}~\citep{yang_pane_2023}; the drug-drug interaction network \textit{OGBL-DDI}~\citep{hu_open_2020}; and the \textit{CoAuthor} network.
We obtained Cora, PubMed, Wikipedia, BlogCatalog, and Facebook from \pyg{} \citep{fey_fast_2019,fey_pyg_2025}, OGBL-DDI from the Open Graph Benchmark~\citep{hu_open_2020}, and for CoAuthor, we used data processed by \citet{tsitsulin_verse_2018}.
Where benchmark splits into training, validation, and test data were available, as for Cora, PubMed, and OGBL-DDI, we used these.
Otherwise, for the remaining node classification datasets, we sampled a split into 70\%/10\%/20\% training/validation/test data.
For CoAuthor, we followed the protocol by \citet{tsitsulin_verse_2018}, treating edges that appeared between the 2014 and 2016 snapshots of the underlying Microsoft Academic Graph as new links.

For synthetic networks, we also created splits based on all edges in the data, with graph reconstruction as downstream tasks in mind.
Here, we used 40\% of all existing edges as training data and split the remaining edges evenly into validation and test data.
However, we ultimately did not evaluate downstream stability on these synthetic datasets, since the resulting predictions were typically either near-random or near-perfect, leaving little meaningful variation to analyze.

\subsection{Embedding Generation}\label{ap:embedding_generation}

\subheader{Embedding Generation.}
\new{
	For all algorithms, tuning was performed at dimension $D=128$, and the selected hyperparameters were subsequently reused across dimensions. This kept the remaining training protocol fixed while varying embedding dimensionality.
	We also briefly investigate the sensitivity of the results compared to tuning at each dimension in Appendix \ref{ap:hyperparameter_sensitivity}.}
Parameter choice was evaluated based on downstream performance, where we used a logistic classifier with standard parameters as implemented in \texttt{scikit-learn} as downstream classifier.
Performance was evaluated based on the average accuracy on five different embeddings computed under a given parameter configuration, using the validation data as described in \Cref{ap:datasets}.
For link prediction and graph reconstruction, we appended the embeddings of each node pair to obtain inputs for the downstream classifier---we note that this differs from the Hadamard product used in the actual evaluation of downstream performance, where we found that this generally performed better.
We did not retune embeddings after changing the downstream protocol, since the purpose of embedding tuning was merely to obtain reasonable baseline-quality representations.

Once best parameters for a combination of dataset and embedding algorithm were determined, these were used throughout all dimensions and training seeds.

\subheader{Implementation Details.}
\new{For ASNE, we used the implementation provided by \texttt{karateclub}~\citep{rozemberczki_karate_2020}. This implementation represents nodes through tokens corresponding to their neighbors and nonzero attributes and learns node embeddings using \emph{Doc2Vec}~\citep{le_distributed_2014}, rather than using the neural architecture described in the original paper.
	For DGI, we used the \texttt{DeepGraphInfomax} class from \texttt{PyTorch Geometric}~\citep{fey_fast_2019,fey_pyg_2025} with GraphSAGE convolutional layers in its encoder. We used an inductive mini-batch setup in which neighborhood sampling was applied during both training and extraction of the final embeddings, rather than the full-graph GCN setup of the reference implementation.
	For GraphSAGE, we used the corresponding class from \texttt{PyTorch Geometric} with mean aggregation. We trained the model through dot-product reconstruction of observed and sampled negative edges, rather than using the random-walk-derived context pairs employed by the reference unsupervised implementation, and applied neighborhood sampling during both training and embedding extraction.
	For node2vec, we used the implementation from \texttt{GRAPE}~\citep{cappelletti_grape_2023}, which delegates both random-walk generation and Skip-Gram training to \texttt{Ensmallen}\footnote{\url{https://github.com/AnacletoLAB/ensmallen}}. This implementation largely follows the original node2vec procedure, using biased second-order random walks and a parallel Skip-Gram optimizer with negative sampling. Random-walk transitions for high-degree nodes are approximated according to the maximum-neighbor setting reported below.
	Finally, for VERSE, we used the reference implementation~\citep{tsitsulin_verse_2018}.}

\subheader{Hyperparameter Choices.}
In the following, we discuss the parameter choices along with the tuning grids used in our optimization protocol.
We start with the parameters that we used as fixed defaults---if parameters are neither mentioned there nor considered in the tuning grids, we used the corresponding defaults from the respective implementations.
\begin{itemize}
	\item \textbf{\textit{ASNE}}: We used a learning rate of 0.05 and trained models for 100 epochs.
	\item \textbf{\textit{DGI}}: Within DGI's encoder, we used GraphSAGE's convolutional layers. Optimization was conducted using an Adam optimizer with a learning rate of 0.001, at a maximum of 200 epochs. Batch size was set to 256. \new{After a minimum of 10 epochs, training was stopped if the loss did not improve by at least 0.1\% relative to its previous best value for 10 consecutive epochs. The final embeddings were extracted from the checkpoint attaining the lowest training loss.}
	Number of sampled neighbors per layer was 10 for the initial two layers and 25 for all subsequent layers. If, in neighbor sampling, the estimated batch size of sampled neighbors exceeded 200,000 (or 100,000 on CoAuthor), these numbers were reduced such that these thresholds would not be surpassed. 
	\item \textbf{\textit{GraphSAGE}}: Optimization was conducted using an Adam optimizer with a learning rate of 0.001 at a maximum of 200 epochs. Batch size was set to 256. \new{We applied the same early-stopping and checkpoint-selection procedure as for DGI and sampled one negative edge per positive edge.}  We sampled 10 neighbors per layer and applied the same reductions as for DGI whenever the estimated number of sampled nodes per batch exceeded 200,000.
	\item \textbf{\textit{node2vec}}: We set the number of walks per node to 10, the number of negative samples per positive to 1, the learning rate to 0.01, and the number of epochs to 50. \new{At each step of a random walk, at most 100 neighbors of the current node were considered when selecting the next node.}
	\item \textbf{\textit{VERSE}}: In VERSE, we used Personalized PageRank to compute similarities between nodes. We set the number of negative samples per positive to 3, and the learning rate to 0.025.
\end{itemize}

\noindent
Finally, we considered the following parameter grids in hyperparameter tuning:
\begin{itemize}
	\item \textbf{\textit{ASNE}:} We varied the down-sampling rate in $\{0.01, 0.001, 0.0001, 0.00001\}$.
	\item \textbf{\textit{DGI}}: We varied number of layers in $\{2,3,4,5\}$.
	\item \textbf{\textit{GraphSAGE}}: We varied number of layers in $\{2,3,4,5\}$.
	\item \textbf{\textit{node2vec}}: We varied walk length in $\{50,80,100\}$, context size in $\{5,10,20\}$, and the walk parameters $p,q\in\{0.5,1,2\}$.
	\item \textbf{\textit{VERSE}}: We varied the personal PageRank restart probability $\alpha\in\{ 0.7, 0.8, 0.85, 0.9 \}$ and the number of negative samples in $\{1,2,3,4,5\}$.
\end{itemize}

\subsection{Downstream Classification}\label{ap:downstream_setup}

\subheader{Implementation Details}.
For both the logistic regression and the MLP classifiers we used the implementations provided by \texttt{scikit-learn}.

\subheader{Experimental Protocol.}
Before generating the predictions used for evaluating downstream performance and functional stability, we tuned the downstream classifiers.
Specifically, for every combination of embedding method, dataset, and embedding dimension, we trained a classifier with a given parameter setting on the training splits of five different embeddings, and used the accuracy on the validation data as the score based on which we determined the optimal parameters.
The final predictions, based on which we assessed downstream performance and functional stability, were then computed on the test splits of each dataset, using the optimal parameters determined in tuning.
We always used the computed embeddings as-is as classifier inputs, except for MLP predictions on DGI embeddings, where we found that normalizing embeddings before training the classifiers substantially improved performance, specifically at higher dimensionality.

\subheader{Hyperparameter Choices.}
When tuning hyperparameters, for the logistic regressor, we varied the regularization weight $C\in\{ 10.0^i: -8\leq i \leq 5\}$, and for the multilayer perceptron, we varied the regularization weight $\alpha\in\{ 10.0^i: -8\leq i \leq 3\}$.

Further, for the MLP classifier, we varied the size of the single inner layer depending on the dimension of its input, aiming to keep model capacity approximately proportional to input dimensionality.
Specifically, we used the following mapping of input dimension $D$ to inner layer size $L$:
\smallskip
\begin{center}
	\begin{tabular}{|l||c|c|c|c|c|c|c|c|c|c|c|c|}
		\hline
		$D$  &  4 & 8 & 16& 32& 64 & 128 & 256 & 512 & 1024 & 2048 & 4096 & 8192\\\hline
		$L$ & 4 & 8 & 12 & 24& 48 & 96 & 128 & 256 & 256 & 256 & 512 & 1024 \\ 
		\hline
	\end{tabular}
\end{center}
\smallskip

Across all experiments, we set the maximum number of iterations for the logistic regressor to 1000, and for the MLP classifier to 500. Other than that, we used the default parameters as provided by \texttt{scikit-learn}.

\FloatBarrier

\newpage
\section{Additional Results}\label{ap:extra_results}

\begin{figure}[b!]
	\centering
	\includegraphics[width=\textwidth]{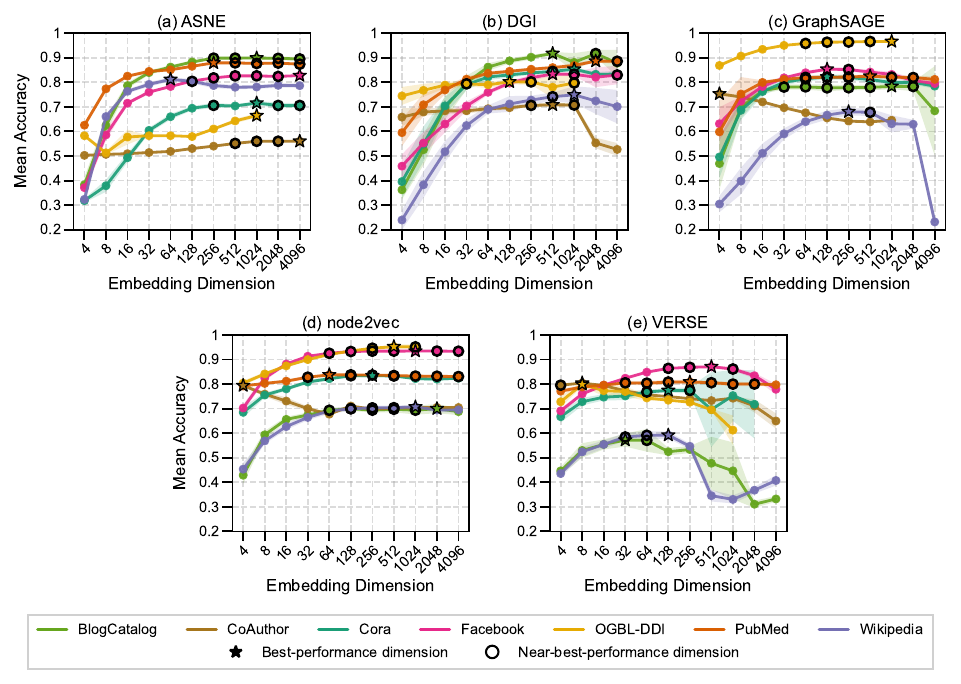}
	\caption{\captionheader{Impact of dimension on downstream performance, using an MLP as downstream classifier.}  We present the average accuracy of each embedding method across 30 training runs using an MLP as downstream classifier. \new{Shaded bands indicate one standard deviation around the respective means.} Stars denote the best-performing dimension for each dataset, dark rings indicate dimensions whose performance is within 0.01 of the optimum.}
	\label{fig:downstream_mlp}
\end{figure}

\subsection{Downstream Performance}\label{ap:downstream_performance}

\Cref{fig:downstream_mlp} shows downstream results when using an MLP as downstream classifier. 
Results on OGBL-DDI for dimensions greater than 1024 are missing due to computational constraints (cf. Appendix \ref{ap:costs}).
The given results are very similar to those using a logistic regressor as downstream classifier, as depicted in \Cref{fig:downstream_logreg}, with performance generally increasing for ASNE and node2vec, while results for GraphSAGE and VERSE often deteriorate at higher dimensions.
For DGI, results often appear to deteriorate as well at high dimensions, though this is more due to higher fluctuation in classification performance across individual runs, as best-performing runs still show increasing performance---it is likely that more focused parameter tuning would stabilize performance at higher accuracy.
Other than that, optimum performance for MLP is often higher than for logistic regression.

\FloatBarrier
\subsection{Representational Stability}\label{ap:representational}

\begin{figure}[b!]
	\centering
	\includegraphics[width=\textwidth]{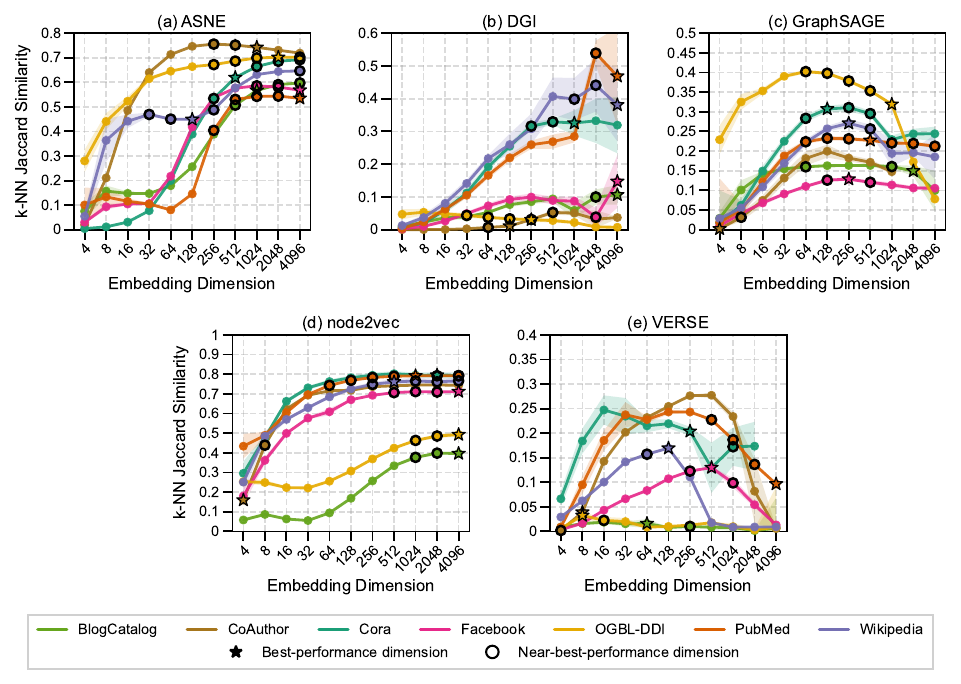}
	\caption{\captionheader{Impact of dimension on representational stability in terms of $k$-NN Jaccard Similarity.} 
		We depict average similarity scores aggregated across all pairs of the 30 computed embeddings (435 pairs in total), where higher values indicate higher stability. \new{Shaded bands indicate one standard deviation around the respective means.} Stars denote the best-performing dimension for each dataset, dark rings indicate dimensions whose performance is within 0.01 of the optimum.
	}
	\label{fig:jaccard}
\end{figure}

\begin{figure}[b!]
	\centering
	\includegraphics[width=\textwidth]{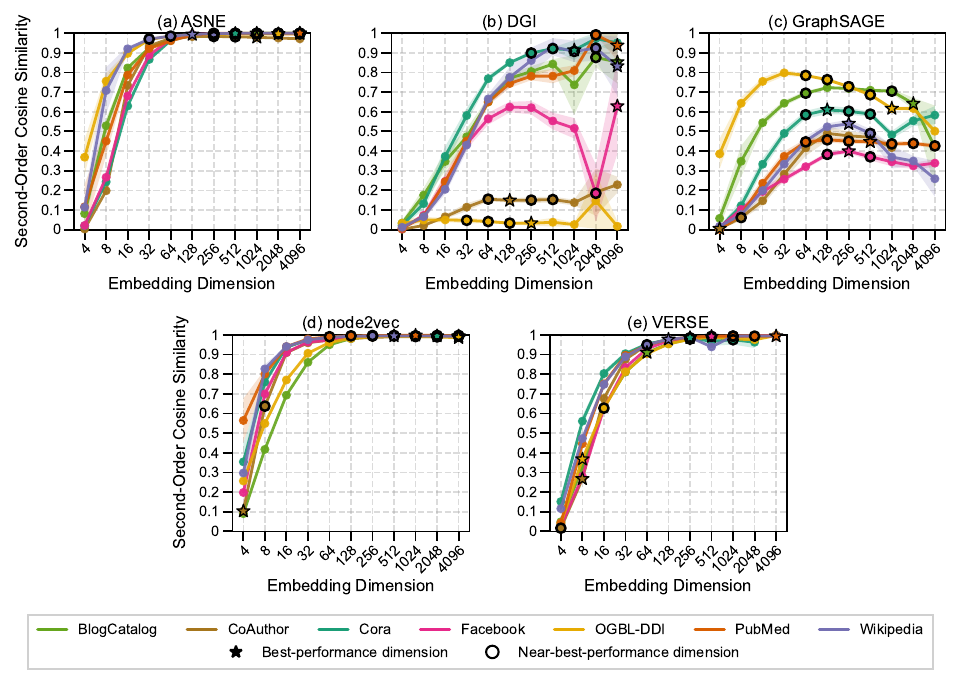}
	\caption{\captionheader{Impact of dimension on representational stability in terms of second-order cosine similarity.} We depict average similarity scores aggregated across all pairs of the 30 computed embeddings (435 pairs in total), where higher values indicate higher stability.
		\new{Shaded bands indicate one standard deviation around the respective means.} 
		Stars indicate the dimension at which the best downstream accuracy was achieved, dark rings indicate dimensions whose performance is within 0.01 of the optimum. Higher values indicate more stable embeddings.}
	\label{fig:2ndcos}
\end{figure}

\begin{figure}[t]
	\centering
	\includegraphics[width=\textwidth]{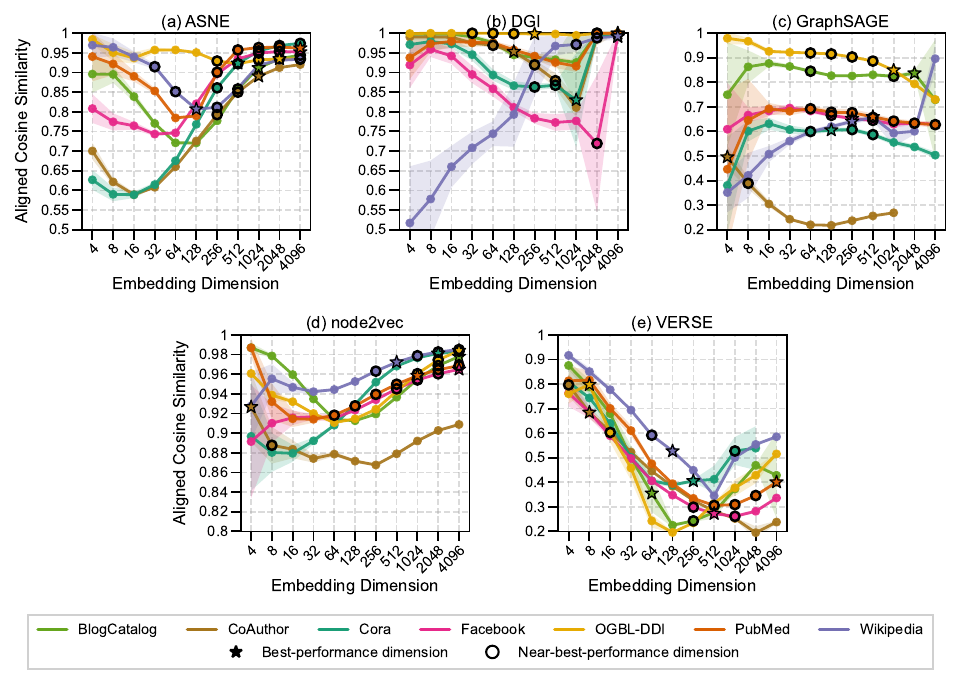}
	\caption{\captionheader{Impact of dimension on representational stability in terms of aligned cosine similarity.} 
		We depict average similarity scores aggregated across all pairs of the 30 computed embeddings (435 pairs in total).
		Higher values indicate higher stability. \new{Shaded bands indicate one standard deviation around the respective means.} 
		Stars denote the best-performing dimension for each dataset, dark rings indicate dimensions whose performance is within 0.01 of the optimum.
	}
	\label{fig:aligncos}
\end{figure}

\begin{figure}[p]
	\centering
	\includegraphics[width=\textwidth]{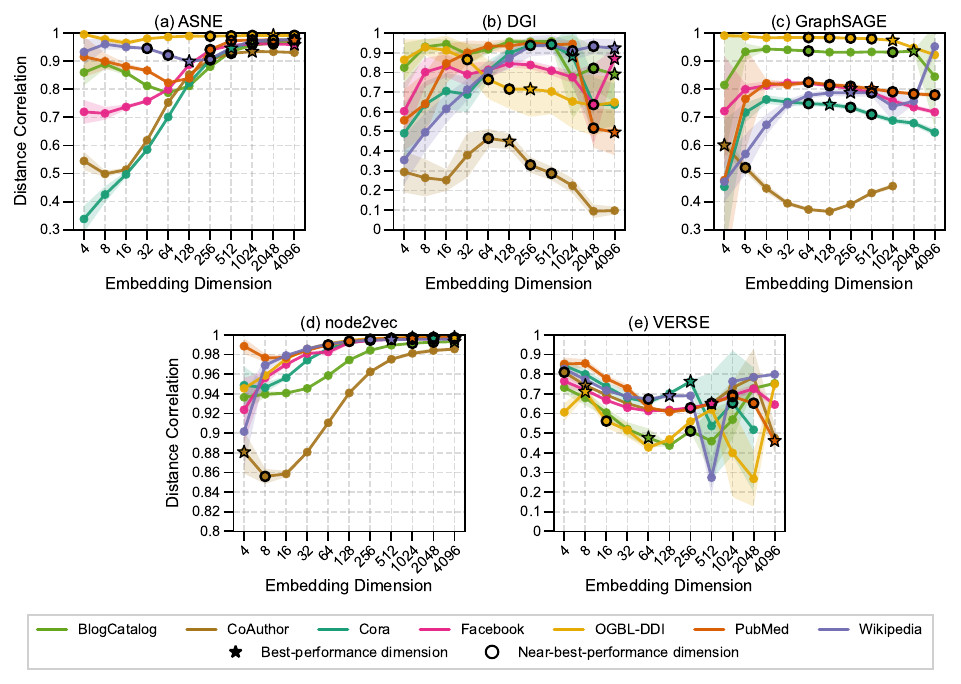}
	\caption{\captionheader{Impact of dimension on representational stability in terms of distance correlation.} \new{Shaded bands indicate one standard deviation around the respective means.} 
		Stars indicate the dimension at which the best downstream accuracy was achieved, dark rings indicate dimensions whose performance is within 0.01 of the optimum. Higher values indicate more stable embeddings.
	}
	\label{fig:distcorr}
\end{figure}

	\Cref{fig:jaccard,fig:2ndcos,fig:aligncos,fig:distcorr} present the complete dataset-specific trajectories underlying the recurring trends summarized in \Cref{tab:summary_repsim}. Overall, the curves broadly follow the depicted method-level patterns, while providing further detail on their behavior across the individual datasets.
	Where partial trends or overall mixed curves are depicted in \Cref{tab:summary_repsim}, these plots also give more context. 
	For DGI, we see that with respect to $k$-NN Jaccard similarity, three curves increase to high levels, three increase at rather low levels, and OGBL-DDI generally remains low with decreasing tendency. With respect to distance correlation, the dataset-dependent curves are even more heterogeneous.
	For GraphSAGE, aligned cosine similarity initially increases and subsequently declines across four datasets. By contrast, CoAuthor and OGBL-DDI lack this initial increase, while Wikipedia instead follows a predominantly increasing trajectory.
	
	The figures also illustrate dataset-specific departures from the consensus trends. While second-order cosine similarity for DGI generally increases toward a high plateau, it remains at substantially lower levels on CoAuthor and particularly OGBL-DDI.
	Similarly, the recurring decrease-and-recovery pattern of ASNE's global stability is only weakly expressed for OGBL-DDI, which remains nearly constant at high levels, and is absent for distance correlation on Cora, which increases across dimensions.
	For DGI, the recurring high-level aligned-cosine trajectory with a later dip is replaced by nearly constant high stability on OGBL-DDI and a nearly linear increase from initially low stability on Wikipedia.

Next to the plain curves of stability across dimensions, the plots also indicate the dimensions yielding optimal and near-optimal downstream performance, respectively, for comparison.
Thereby we observe that, for example, for ASNE on Cora and Wikipedia, near-optimal performance extends across intermediate and high dimensions, while $k$-NN Jaccard and aligned cosine similarity continue increasing toward their high-dimensional levels. These trajectories illustrate how the stability maximum may fall within a broad performance plateau without occurring at the strictly best-performing dimension.

\clearpage
\FloatBarrier
\subsection{Functional Stability}\label{ap:funcsim_logreg}

\begin{figure}[b!]
	\centering
	\includegraphics[width=\textwidth]{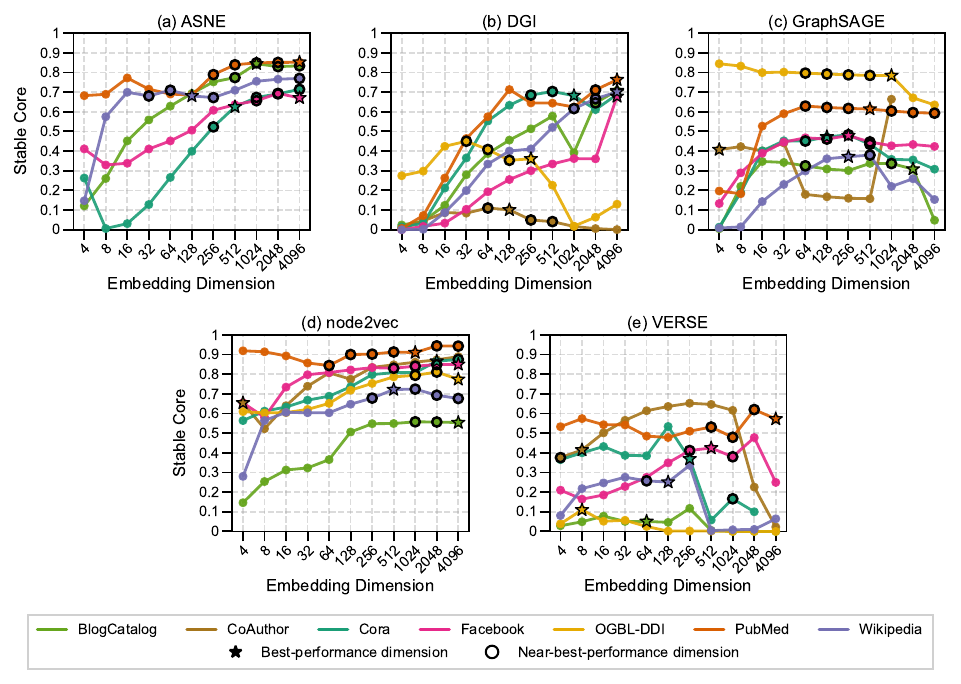}
	\caption{\captionheader{Impact of dimension on functional stability in terms of stable core.}
		The stable core corresponds to the fraction of test instances for which predictions across all 30 embeddings agree. Higher values indicate more consistent predictions. 
		Stars denote the best-performing dimension for each dataset, dark rings indicate dimensions whose performance is within 0.01 of the optimum.
		Stable predictions are often achieved at higher dimensions, particularly for ASNE, DGI, and node2vec. By contrast, functional stability for GraphSAGE and VERSE frequently decreases again at high dimensions.
	}
	\label{fig:stablecore}
\end{figure}

\begin{figure}[t]
	\centering
	\includegraphics[width=\textwidth]{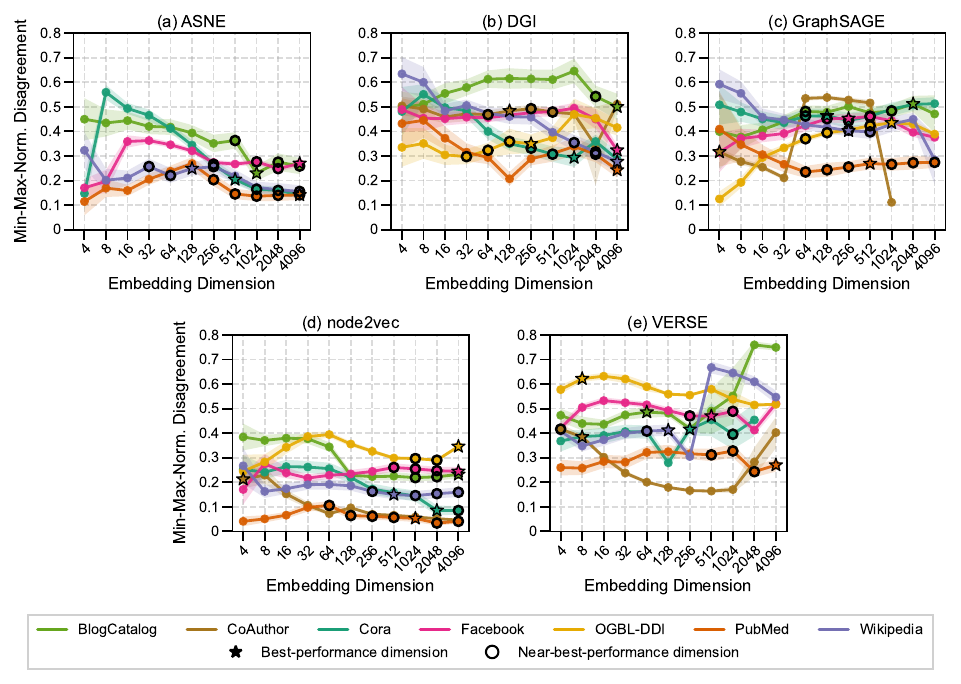}
	\caption{\captionheader{Impact of dimension on functional stability in terms of min-max normalized disagreement.}
		We depict average disagreement scores aggregated across all pairs of the 30 computed embeddings (435 pairs in total).
		\new{Shaded bands indicate one standard deviation around the respective means.} Stars indicate the dimension at which the best downstream accuracy was achieved, dark rings indicate dimensions whose performance is within 0.01 of the optimum. 
		Lower values indicate more stable downstream predictions.
	}
	\label{fig:normdis}
\end{figure}

\begin{figure}[b]
	\centering
	\includegraphics[width=\textwidth]{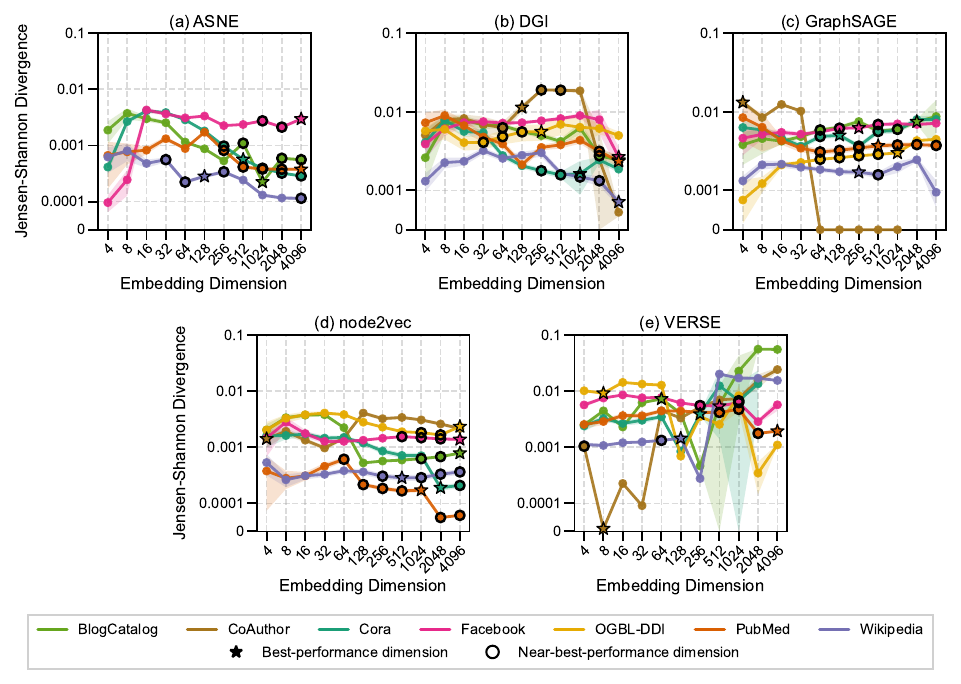}
	\caption{\captionheader{Impact of dimension on functional stability in terms of Jensen-Shannon divergence.}  
		We depict average divergence values aggregated across all pairs of the 30 computed embeddings (435 pairs in total).
		Lower values indicate higher similarity.
		\new{Shaded bands indicate one standard deviation around the respective means.} Stars denote the best-performing dimension for each dataset, dark rings indicate dimensions whose performance is within 0.01 of the optimum.
	}
	\label{fig:jsd}
\end{figure}

\begin{revisionblock}

Similar to the previous subsection on representational similarity results, \Cref{fig:stablecore,fig:disagreement,fig:normdis,fig:jsd} present the detailed results including the dataset-specific trajectories underlying the functional-stability trends summarized in \Cref{tab:summary_funcsim}, where a logistic regressor was used as downstream classifier.

As discussed in the main results, stable core and disagreement generally exhibit corresponding trajectories, with increases in stable core accompanied by decreases in disagreement.
Yet we observe that deviations from the consensus are often tied to the link prediction datasets, CoAuthor and OGBL-DDI. 
This is observed for DGI, where both datasets clearly depart from the recurring improvement toward higher dimensions, and for GraphSAGE, where they lack the initial improvement followed by a moderate high-dimensional decline observed across the remaining datasets.
The corresponding ASNE results are not shown because its embeddings provided little useful signal for either link prediction task (cf. \Cref{sec:performance}).
For VERSE, the partial increase-and-decline trend in stable core is absent on BlogCatalog, OGBL-DDI, and PubMed, while the recurring intermediate-dimensional minimum in disagreement is absent on OGBL-DDI and PubMed.

	\begin{figure}[b!]
		\centering
		\includegraphics[width=\textwidth]{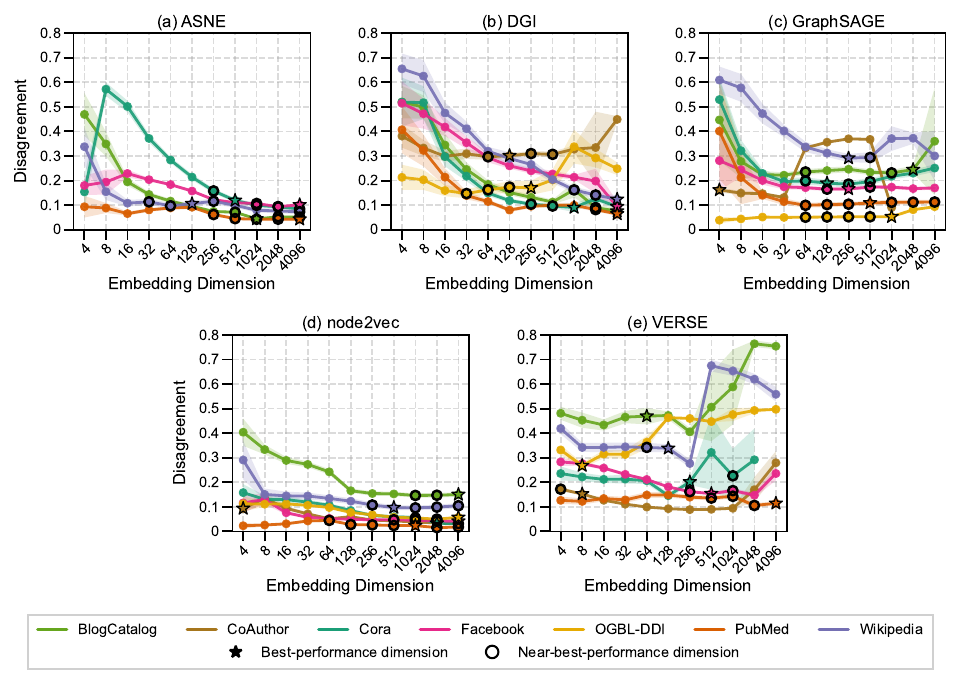}
		\caption{\captionheader{Impact of dimension on functional stability in terms of disagreement.} 
			We depict average disagreement aggregated across all pairs of the 30 computed embeddings (435 pairs in total).
			\new{Shaded bands indicate one standard deviation around the respective means.} 
			Stars indicate the dimension at which the best downstream accuracy was achieved, dark rings indicate dimensions whose performance is within 0.01 of the optimum.
			Lower values indicate more stable downstream predictions.
		}
		\label{fig:disagreement}
	\end{figure}

The figures also provide further context for the partial and mixed trends obtained with the remaining measures.
For ASNE, the partial min-max-normalized disagreement trend is supported by Cora, Facebook, and PubMed, which initially increase in disagreement before declining again. BlogCatalog instead declines from the lowest dimensions onward, while Wikipedia exhibits only moderate fluctuations.
For the remaining embedding methods, min-max-normalized disagreement shows no recurring trajectory across datasets.
Similarly, JSD remains at generally very low levels, commonly between $10^{-4}$ and $10^{-2}$, but exhibits no recurring dimensionality-dependent trajectory.

Again, these figures also point out dimensions yielding optimal and near-optimal downstream performance, respectively.
This yields additional examples of how stability can continue changing within a broad region of near-optimal performance.
One such example is ASNE on Cora, where the stable core continues increasing and disagreement continues decreasing across dimensions already yielding near-optimal performance.

\end{revisionblock}

\clearpage
\FloatBarrier
\subsection{Functional Stability with MLP Classifier}\label{ap:funcsim_mlp}

\begin{figure}[b!]
	\centering
	\includegraphics[width=\textwidth]{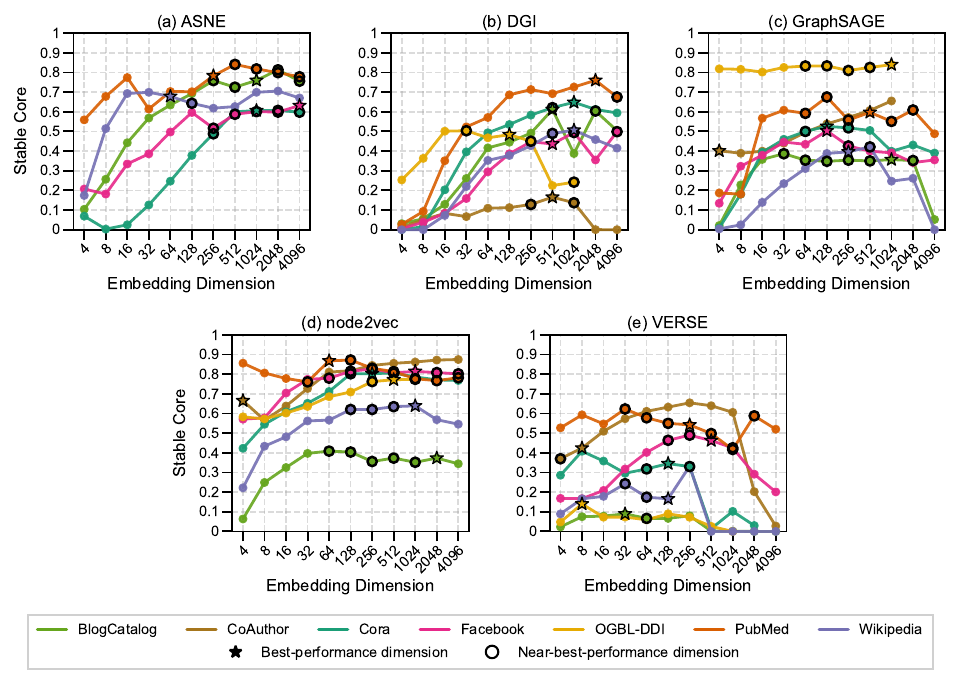}
	\caption{\captionheader{Impact of dimension on functional stability in terms of stable core, using an MLP as downstream classifier.} 
		Stars indicate the dimension at which the best downstream accuracy was achieved (cf. \Cref{fig:downstream_mlp}), dark rings indicate dimensions whose performance is within 0.01 of the optimum.
		Higher values indicate more stable downstream predictions.
	}
	\label{fig:stablecore_mlp}
\end{figure}

\begin{figure}[t]
	\centering
	\includegraphics[width=\textwidth]{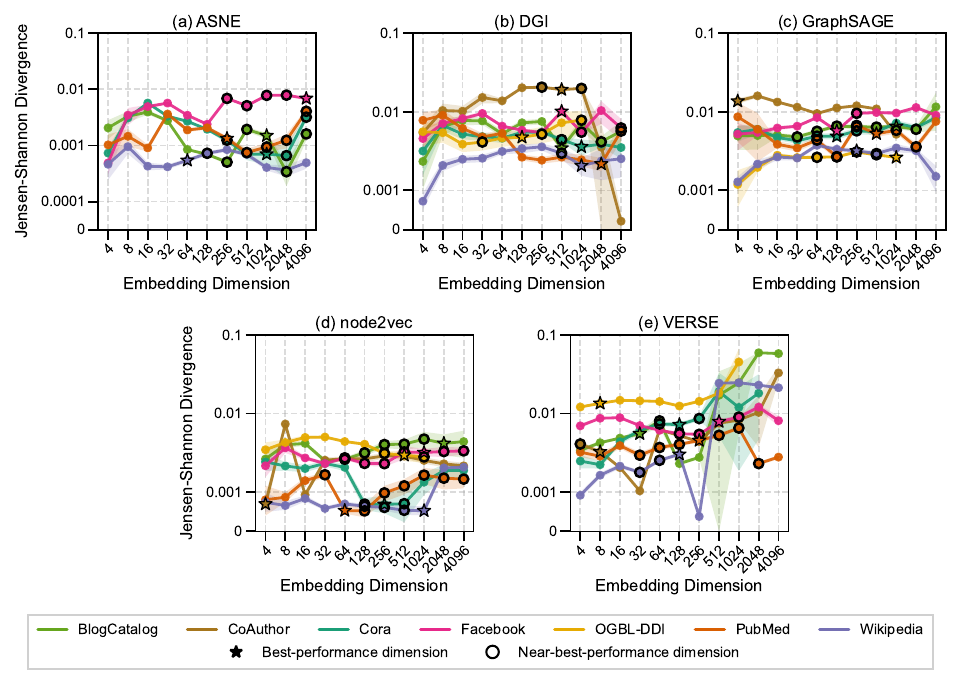}
	\caption{\captionheader{Impact of dimension on functional stability in terms of Jensen-Shannon divergence, using an MLP as downstream classifier.}
		We depict average divergence scores aggregated across all pairs of the 30 computed embeddings (435 pairs in total).
		\new{Shaded bands indicate one standard deviation around the respective means.} 
		Stars indicate the dimension at which the best downstream accuracy was achieved (cf. \Cref{fig:downstream_mlp}), dark rings indicate dimensions whose performance is within 0.01 of the optimum.
		Lower values indicate more stable downstream predictions.
	}
	\label{fig:jsd_mlp}
\end{figure}

\begin{figure}[t]
	\centering
	\includegraphics[width=\textwidth]{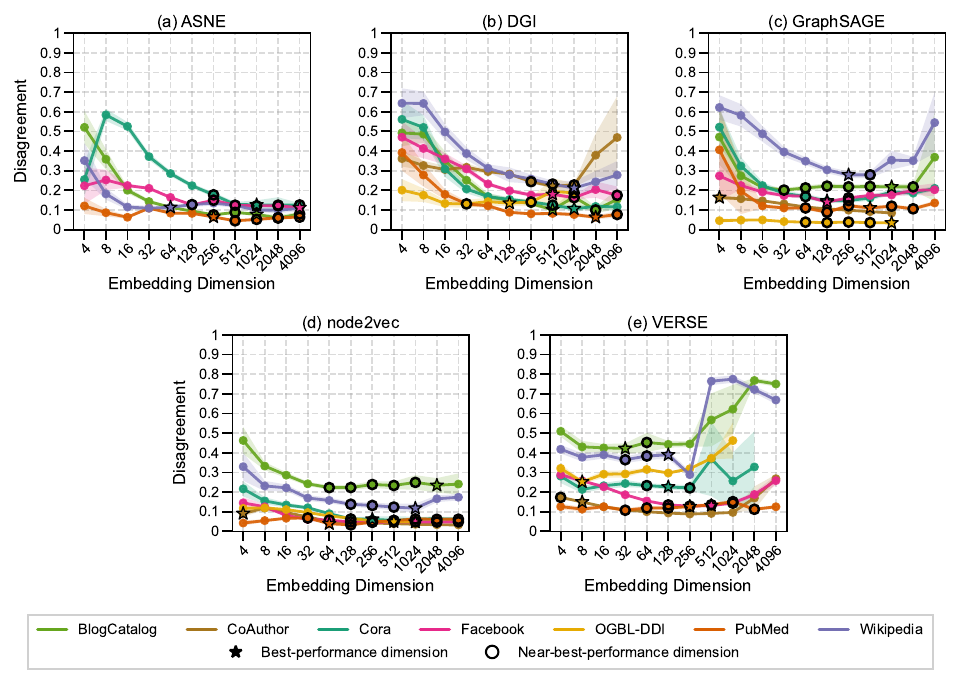}
	\caption{\captionheader{Impact of dimension on functional stability in terms of disagreement, using an MLP as downstream classifier.}
		We depict average disagreement scores aggregated across all pairs of the 30 computed embeddings (435 pairs in total).
		\new{Shaded bands indicate one standard deviation around the respective means.} Stars indicate the dimension at which the best downstream accuracy was achieved (cf. \Cref{fig:downstream_mlp}), dark rings indicate dimensions whose performance is within 0.01 of the optimum.
		Lower values indicate more stable downstream predictions.
	}
	\label{fig:disagreement_mlp}
\end{figure}

\begin{figure}[t]
	\centering
	\includegraphics[width=\textwidth]{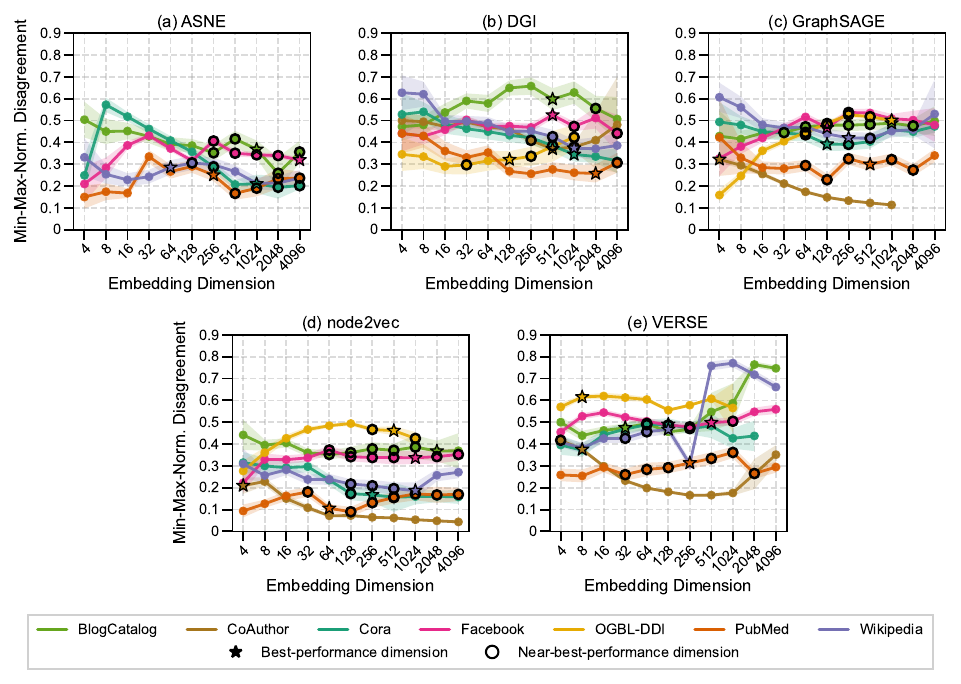}
	\caption{\captionheader{Impact of dimension on functional stability in terms of min-max normalized disagreement, using an MLP as downstream classifier.}
		We depict average disagreement scores aggregated across all pairs of the 30 computed embeddings (435 pairs in total).
		\new{Shaded bands indicate one standard deviation around the respective means.} Stars indicate the dimension at which the best downstream accuracy was achieved (cf. \Cref{fig:downstream_mlp}), dark rings indicate dimensions whose performance is within 0.01 of the optimum.
		Lower values indicate more stable downstream predictions.
	}
	\label{fig:normdis_mlp}
\end{figure}

Figures \Cref{fig:stablecore_mlp,fig:disagreement_mlp,fig:normdis_mlp,fig:jsd_mlp} provide functional stability results obtained using a multilayer perceptron (MLP) as downstream classifier. 
Results on OGBL-DDI for dimensions greater than 1024 are missing due to computational constraints.
Overall, the results largely agree with those obtained using logistic regression and support the same qualitative conclusions. However, instability at larger embedding dimensions is occasionally more pronounced, reflecting the additional stochasticity introduced by the downstream model itself.

\FloatBarrier

\begin{figure}[b!]
	\centering
	\includegraphics[width=\textwidth]{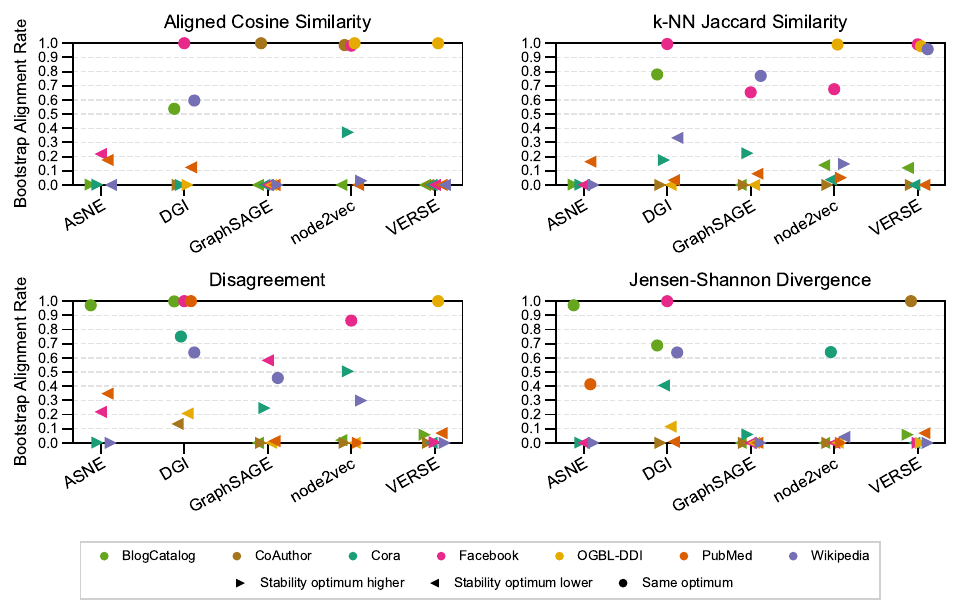}
	\caption{
		\new{\captionheader{Bootstrap alignment of stability-optimal dimensions with dimensions yielding optimal performance.}
			Each point shows, for one embedding method, stability measure, and dataset, the fraction of 10,000 bootstrap replicates in which the stability-optimal dimension is among the dimensions attaining the replicate-specific maximum downstream accuracy.
			Colors identify datasets.
			Right- and left-pointing triangles indicate whether, using all runs, the stability optimum occurs at a higher or lower dimension than the performance optimum, respectively. Circles indicate coinciding optima.
	}}
	\label{fig:bootstrap_alignment_strict}
\end{figure}

\subsection{Alignment Between Optimal Dimensions for Stability and Performance}\label{ap:bootstrap}

\Cref{fig:bootstrap_alignment_strict} complements the threshold-based results presented in \Cref{fig:bootstrap_alignment} by requiring exact agreement between the stability- and performance-optimal dimensions.
Under this stricter criterion, bootstrap alignment rates decrease substantially across many configurations.
In particular, where the optima estimated from the complete set of runs occur at different dimensions, strict alignment rates are almost always below 0.5, and close to zero in the majority of cases.
These results indicate that the observed differences between the exact optima usually persist across bootstrap replicates, even where the stability-optimal dimension frequently belongs to the broader set of dimensions yielding near-optimal performance.

\begin{figure}[b]
	\centering
	\includegraphics[width=\textwidth]{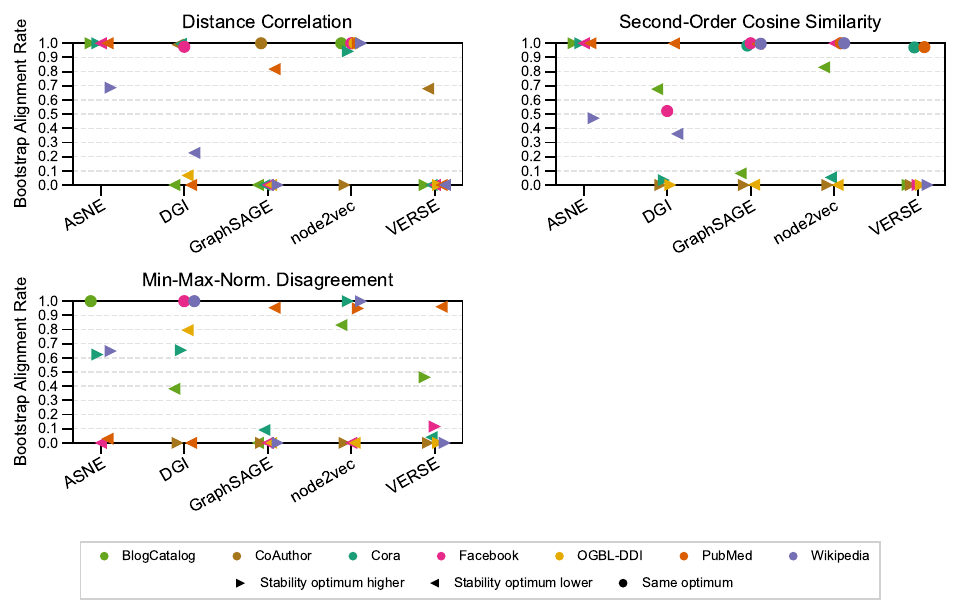}
	\caption{
		\new{\captionheader{Bootstrap alignment of stability-optimal dimensions with dimensions yielding near-optimal performance.}
			Each point shows, for one embedding method, stability measure, and dataset, the fraction of 10,000 bootstrap replicates in which the stability-optimal dimension is among the dimensions yielding downstream accuracy within 0.01 of the replicate-specific maximum.
			Colors identify datasets.
			Right- and left-pointing triangles indicate whether, using all runs, the stability optimum occurs at a higher or lower dimension than the performance optimum, respectively. Circles indicate coinciding optima.
	}}
	\label{fig:bootstrap_alignment_ap}
\end{figure}

\begin{figure}[b]
	\centering
	\includegraphics[width=\textwidth]{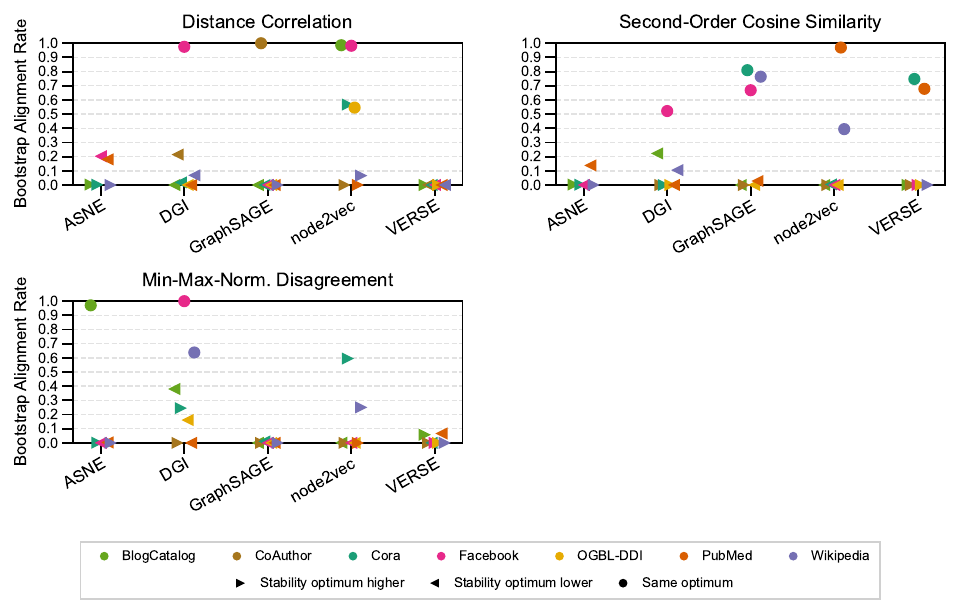}
	\caption{
		\new{\captionheader{Bootstrap alignment of stability-optimal dimensions with dimensions yielding optimal performance.}
			Each point shows, for one embedding method, stability measure, and dataset, the fraction of 10,000 bootstrap replicates in which the stability-optimal dimension is among the dimensions attaining the replicate-specific maximum downstream accuracy.
			Colors identify datasets.
			Right- and left-pointing triangles indicate whether, using all runs, the stability optimum occurs at a higher or lower dimension than the performance optimum, respectively. Circles indicate coinciding optima.
	}}
	\label{fig:bootstrap_alignment_strict_ap}
\end{figure}

\Cref{fig:bootstrap_alignment_ap,fig:bootstrap_alignment_strict_ap} present the corresponding threshold-based and strict-alignment results for distance correlation, second-order cosine similarity, and min-max-normalized disagreement.
Stable core was not included in this analysis because it is defined jointly across all runs: duplicates introduced through bootstrap resampling reduce the effective number of distinct predictions and would thereby mechanically inflate the estimated stable core.

For distance correlation and second-order cosine similarity, alignment again varies substantially across datasets and embedding methods, without a recurring pattern that would conflict with the findings presented in the main text.
Min-max-normalized disagreement likewise exhibits mixed alignment and therefore does not reproduce the comparatively frequent alignment observed for unnormalized disagreement.
As for the measures presented in the main text, applying the strict criterion generally produces lower alignment rates.

\FloatBarrier
\subsection{Impact of Network Size}\label{ap:size}

\begin{figure}[h!]
	\centering
	\includegraphics[width=\textwidth]{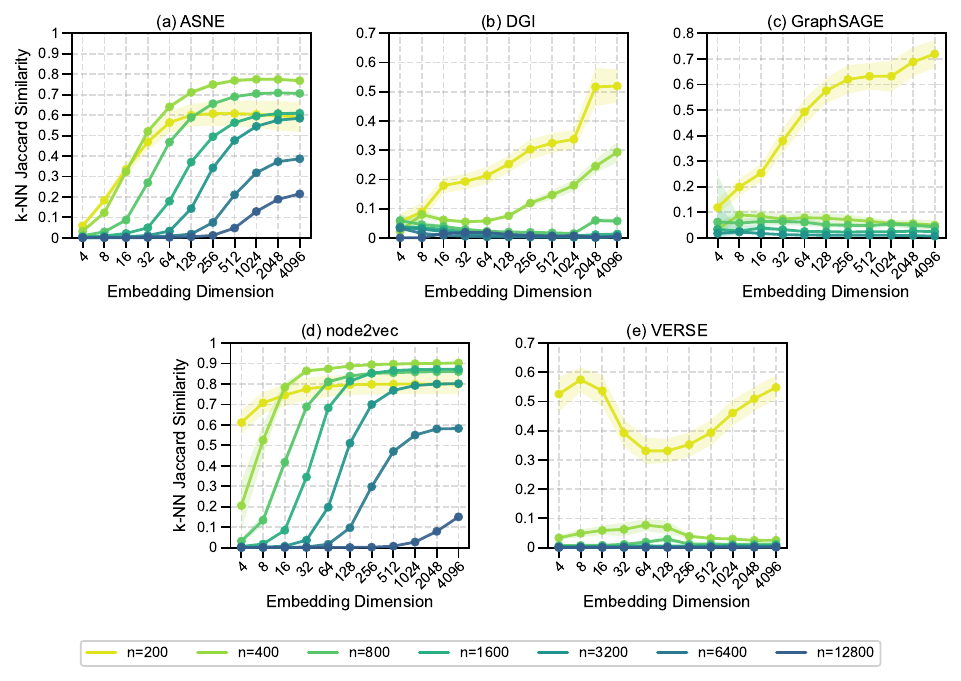}
	\caption{\captionheader{Interplay between network size and embedding dimension on Barabasi-Albert graphs in terms of $k$-NN Jaccard similarity.}
		We depict $k$-NN Jaccard similarity averaged over all pairs of 10 embeddings of the same graphs, taken across 10 different instantiations of the Barabasi-Albert graph at density $d=0.01$, with varying size as indicated by line color. \new{Shaded bands indicate one standard deviation around the respective means.} 
		Higher values indicate more stable embeddings.
	}
	\label{fig:ba_size_jaccard}
\end{figure}

\begin{figure}[t]
	\centering
	\includegraphics[width=\textwidth]{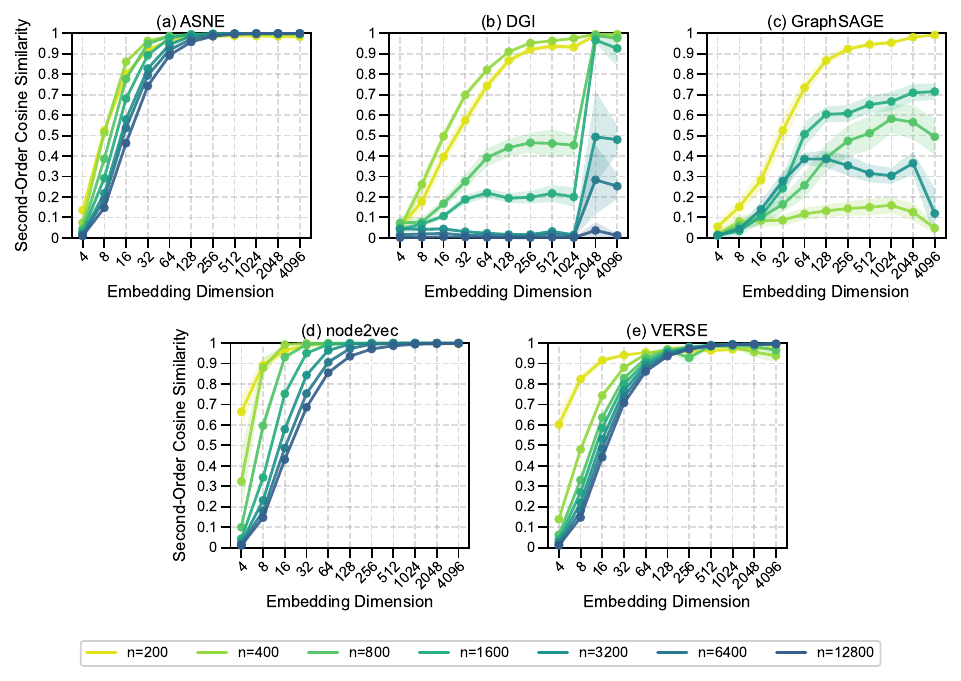}
	\caption{\captionheader{Interplay between network size and embedding dimension on Barabasi-Albert graphs in terms of second-order cosine similarity.}
		We depict second-order cosine similarity averaged over all pairs of 10 embeddings of the same graphs, taken across 10 different instantiations of the Barabasi-Albert graph at density $d=0.01$, with varying size as indicated by line color. \new{Shaded bands indicate one standard deviation around the respective means.} 
		Higher values indicate more stable embeddings.
	}
	\label{fig:ba_size_2ndcos}
\end{figure}

\begin{figure}[t]
	\centering
	\includegraphics[width=\textwidth]{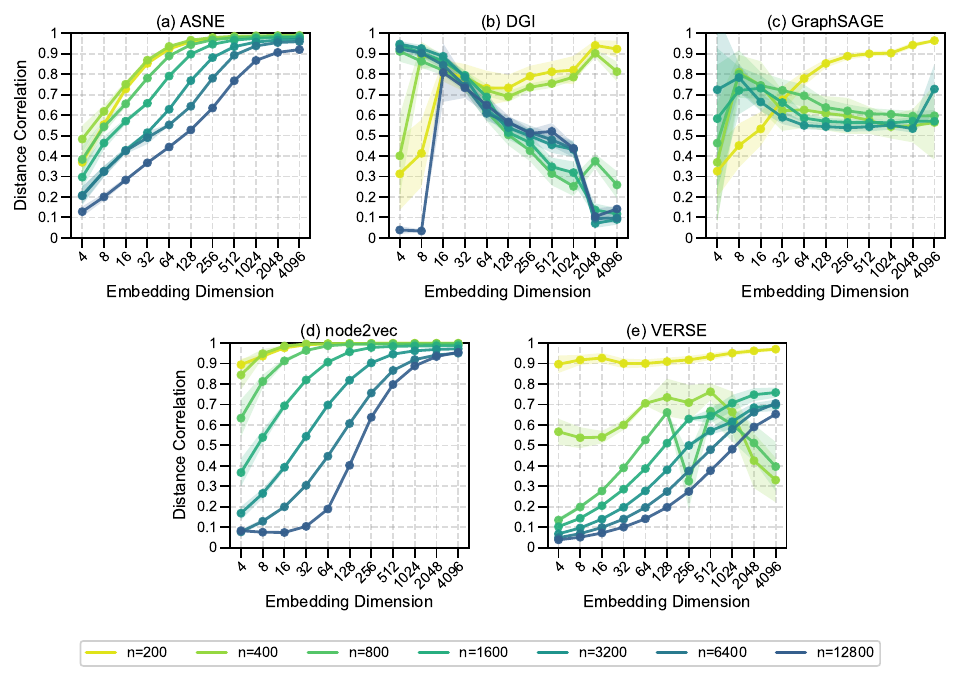}
	\caption{\captionheader{Interplay between network size and embedding dimension on Barabasi-Albert graphs in terms of distance correlation.}
		We depict distance correlation averaged over all pairs of 10 embeddings of the same graphs, taken across 10 different instantiations of the Barabasi-Albert graph at density $d=0.01$, with varying size as indicated by line color. \new{Shaded bands indicate one standard deviation around the respective means.} 
		Higher values indicate more stable embeddings.
	}
	\label{fig:ba_size_distcorr}
\end{figure}

\begin{figure}[t]
	\centering
	\includegraphics[width=\textwidth]{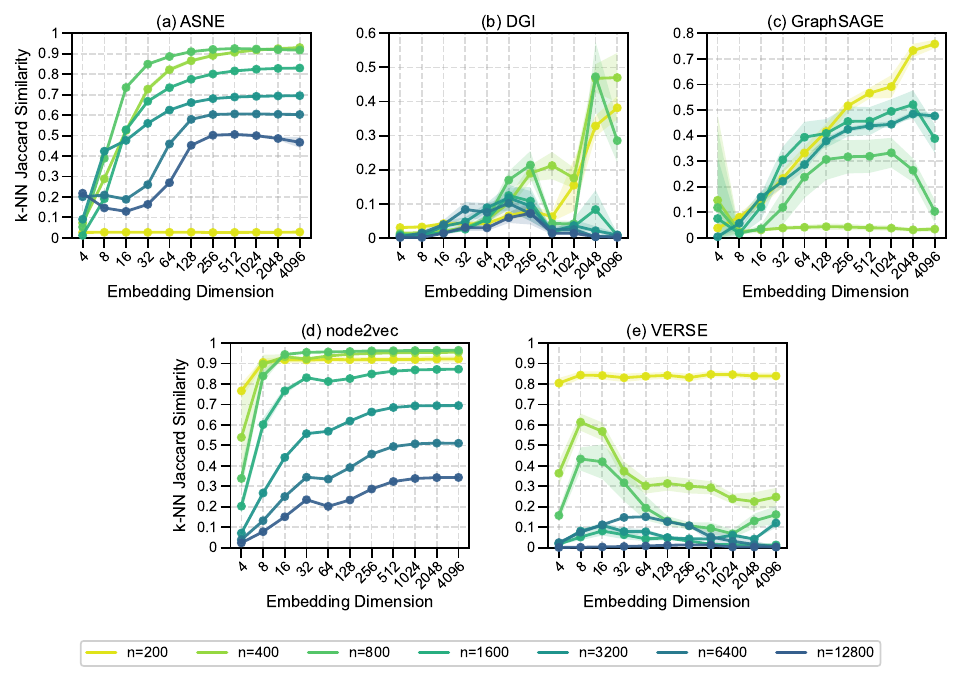}
	\caption{\captionheader{Interplay between network size and embedding dimension on Watts-Strogatz graphs in terms of $k$-NN Jaccard similarity.}
		We depict $k$-NN Jaccard similarity averaged over all pairs of 10 embeddings of the same graphs, taken across 10 different instantiations of the Watts-Strogatz graph at density $d=0.01$, with varying size as indicated by line color. \new{Shaded bands indicate one standard deviation around the respective means.} 
		Higher values indicate more stable embeddings.
	}
	\label{fig:ws_size_jaccard}
\end{figure}

\begin{figure}[t]
	\centering
	\includegraphics[width=\textwidth]{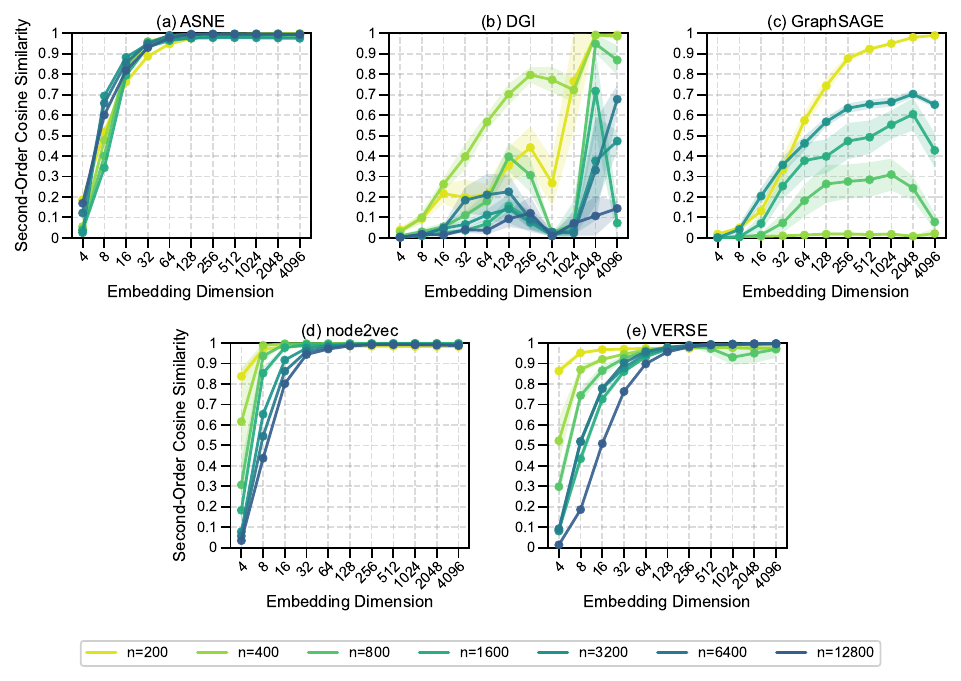}
	\caption{\captionheader{Interplay between network size and embedding dimension on Watts-Strogatz graphs in terms of second-order cosine similarity.}
		We depict second-order cosine similarity averaged over all pairs of 10 embeddings of the same graphs, taken across 10 different instantiations of the Watts-Strogatz graph at density $d=0.01$, with varying size as indicated by line color. \new{Shaded bands indicate one standard deviation around the respective means.} 
		Higher values indicate more stable embeddings.
	}
	\label{fig:ws_size_2ndcos}
\end{figure}

\begin{figure}[t]
	\centering
	\includegraphics[width=\textwidth]{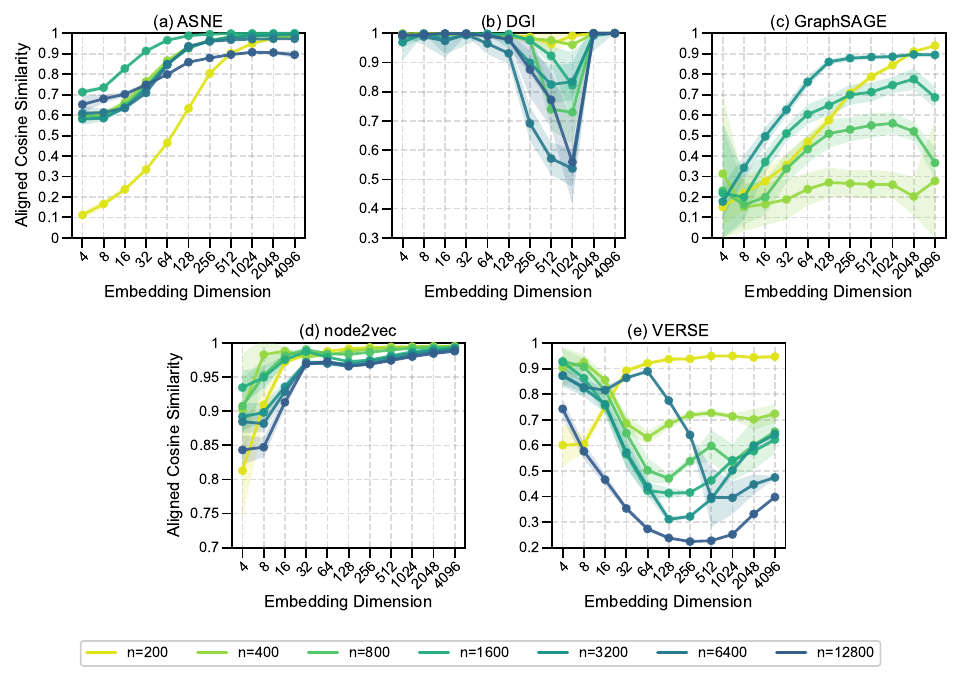}
	\caption{\captionheader{Interplay between network size and embedding dimension on Watts-Strogatz graphs in terms of aligned cosine similarity.}
		We depict aligned cosine similarity averaged over all pairs of 10 embeddings of the same graphs, taken across 10 different instantiations of the Watts-Strogatz graph at density $d=0.01$, with varying size as indicated by line color. \new{Shaded bands indicate one standard deviation around the respective means.} 
		Higher values indicate more stable embeddings.
	}
	\label{fig:ws_size_aligned_cossim}
\end{figure}

\begin{figure}[t]
	\centering
	\includegraphics[width=\textwidth]{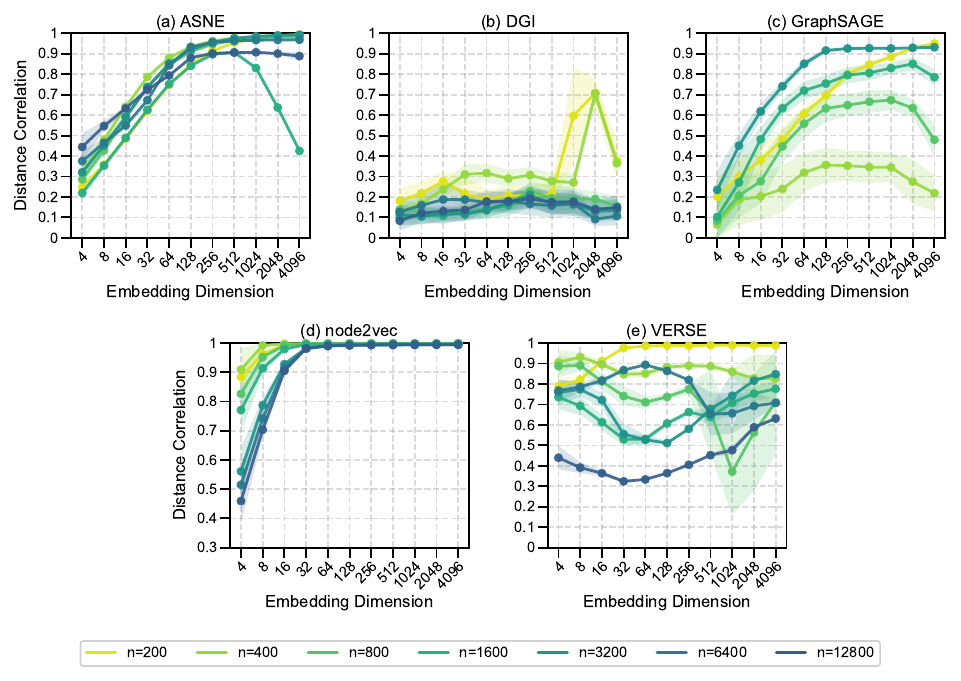}
	\caption{\captionheader{Interplay between network size and embedding dimension on Watts-Strogatz graphs in terms of distance correlation.}
		We depict distance correlation averaged over all pairs of 10 embeddings of the same graphs, taken across 10 different instantiations of the Watts-Strogatz graph at density $d=0.01$, with varying size as indicated by line color. \new{Shaded bands indicate one standard deviation around the respective means.} 
		Higher values indicate more stable embeddings.
	}
	\label{fig:ws_size_distcorr}
\end{figure}

Figures \ref{fig:ba_size_jaccard}--\ref{fig:ws_size_distcorr} provide the complete results for the interaction between graph size and embedding dimensionality across both graph models and all representational similarity measures.
Overall, the trends are broadly consistent with those discussed in the main text. In particular, node2vec and ASNE frequently require larger dimensions before stable high-dimensional regimes emerge as graph size increases. 
Similar tendencies are often visible for VERSE, whereas DGI and GraphSAGE remain less consistent. While the exact shape and strength of these effects depend on the similarity measure considered, graph size generally exhibits more systematic relationships with stability than graph density.

\FloatBarrier

\subsection{Impact of Network Density}\label{ap:density}

\begin{figure}[h!]
	\centering
	\includegraphics[width=\textwidth]{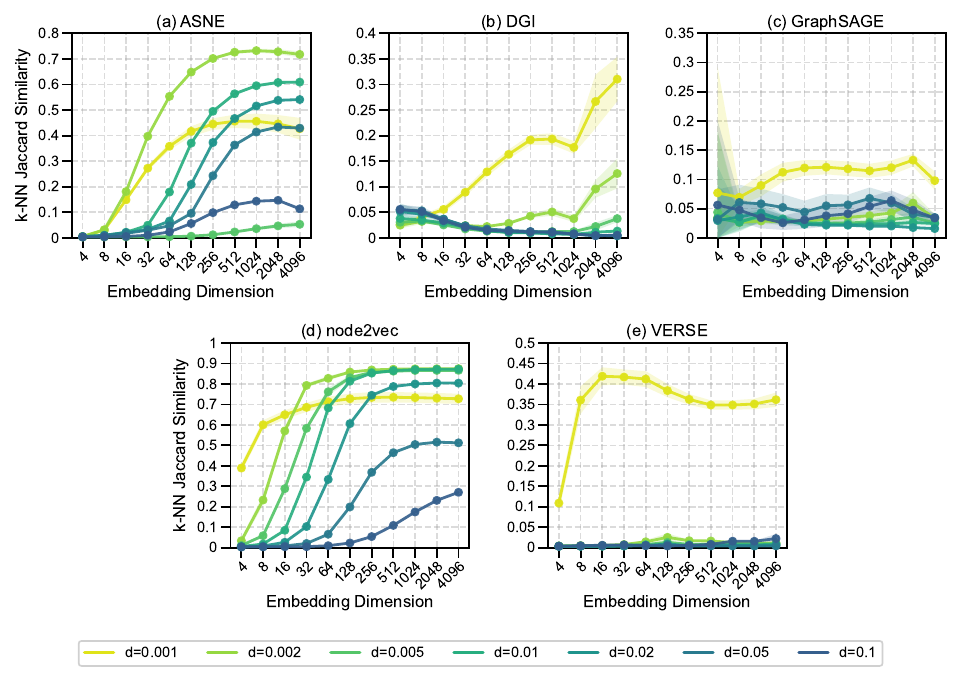}
	\caption{\captionheader{Interplay between network density and embedding dimension on Barabasi-Albert graphs in terms of $k$-NN Jaccard similarity.}
		We depict $k$-NN Jaccard similarity averaged across all pairs of 10 embeddings computed on the same graph, aggregated over 10 independently generated Barabasi-Albert graphs with $N=1600$ nodes and varying density as indicated by line color. \new{Shaded bands indicate one standard deviation around the respective means.} 
		Higher values indicate more stable embeddings.
	}
	\label{fig:ba_density_jaccard}
\end{figure}

\begin{figure}[t]
	\centering
	\includegraphics[width=\textwidth]{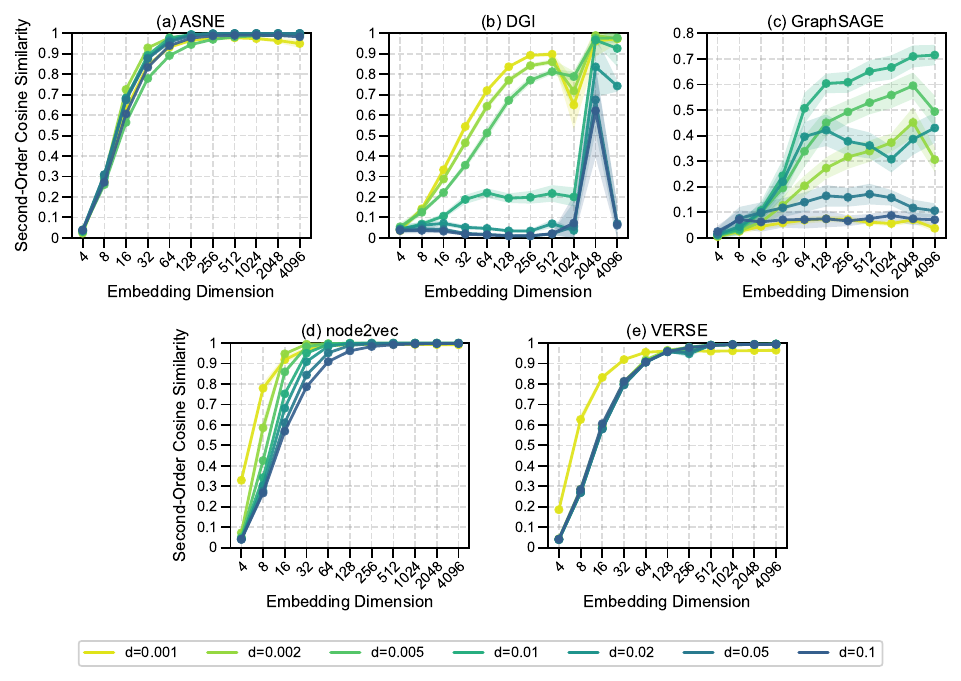}
	\caption{\captionheader{Interplay between network density and embedding dimension on Barabasi-Albert graphs in terms of second-order cosine similarity.}
		We depict second-order cosine similarity averaged across all pairs of 10 embeddings computed on the same graph, aggregated over 10 independently generated Barabasi-Albert graphs with $N=1600$ nodes and varying density as indicated by line color. \new{Shaded bands indicate one standard deviation around the respective means.} 
		Higher values indicate more stable embeddings.
	}
	\label{fig:ba_density_2ndcos}
\end{figure}

\begin{figure}[t]
	\centering
	\includegraphics[width=\textwidth]{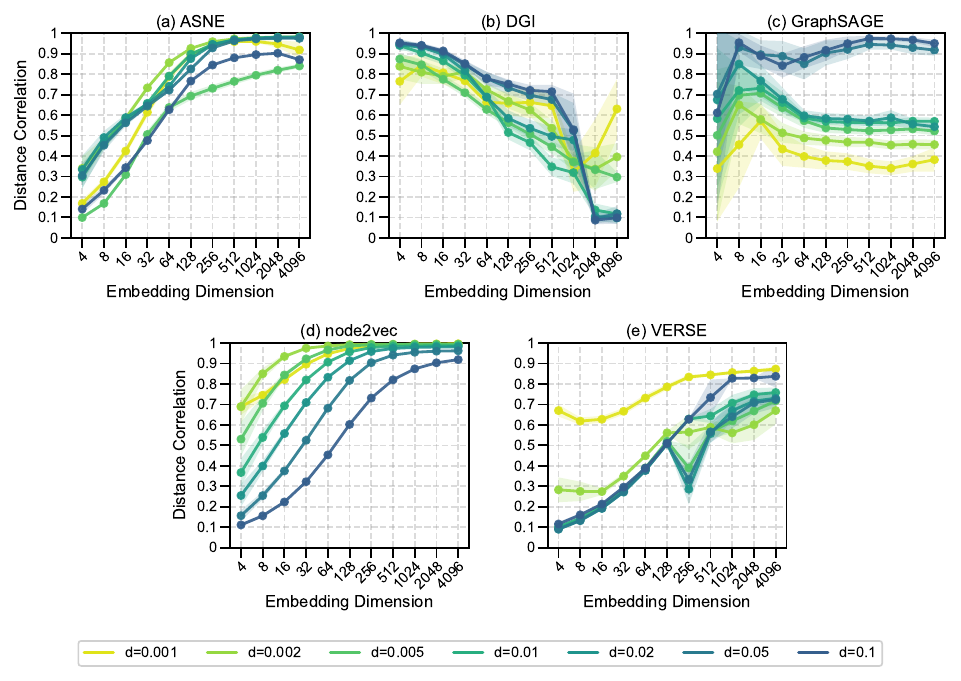}
	\caption{\captionheader{Interplay between network density and embedding dimension on Barabasi-Albert graphs in terms of distance correlation.}
		We depict distance correlation averaged across all pairs of 10 embeddings computed on the same graph, aggregated over 10 independently generated Barabasi-Albert graphs with $N=1600$ nodes and varying density as indicated by line color. \new{Shaded bands indicate one standard deviation around the respective means.} 
		Higher values indicate more stable embeddings.
	}
	\label{fig:ba_density_distcorr}
\end{figure}

\begin{figure}[t]
	\centering
	\includegraphics[width=\textwidth]{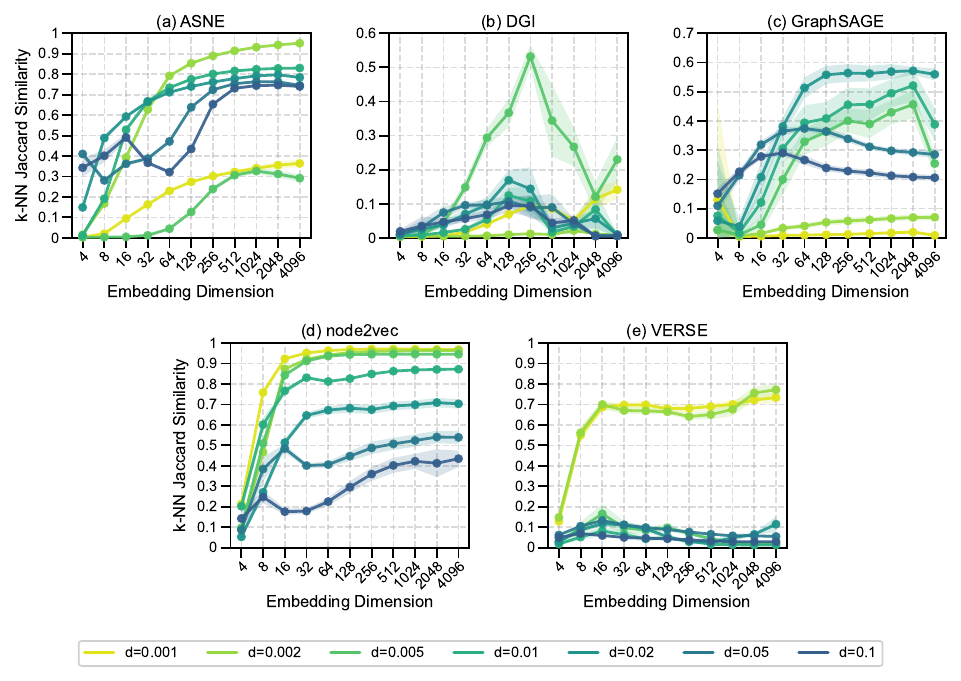}
	\caption{\captionheader{Interplay between network density and embedding dimension on Watts-Strogatz graphs in terms of $k$-NN Jaccard similarity.}
		We depict $k$-NN Jaccard similarity averaged across all pairs of 10 embeddings computed on the same graph, aggregated over 10 independently generated Watts-Strogatz graphs with $N=1600$ nodes and varying density as indicated by line color. \new{Shaded bands indicate one standard deviation around the respective means.} 
		Higher values indicate more stable embeddings.
	}
	\label{fig:ws_density_jaccard}
\end{figure}

\begin{figure}[t]
	\centering
	\includegraphics[width=\textwidth]{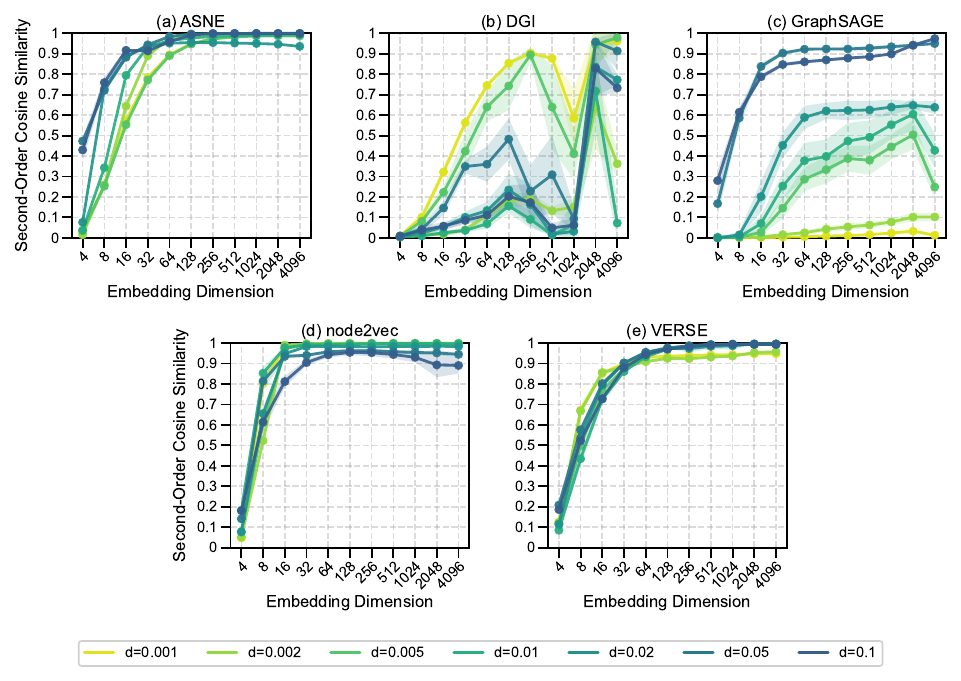}
	\caption{\captionheader{Interplay between network density and embedding dimension on Watts-Strogatz graphs in terms of second-order cosine similarity.}
		We depict second-order cosine similarity averaged across all pairs of 10 embeddings computed on the same graph, aggregated over 10 independently generated Watts-Strogatz graphs with $N=1600$ nodes and varying density as indicated by line color. \new{Shaded bands indicate one standard deviation around the respective means.} 
		Higher values indicate more stable embeddings.
	}
	\label{fig:ws_density_2ndcos}
\end{figure}

\begin{figure}[t]
	\centering
	\includegraphics[width=\textwidth]{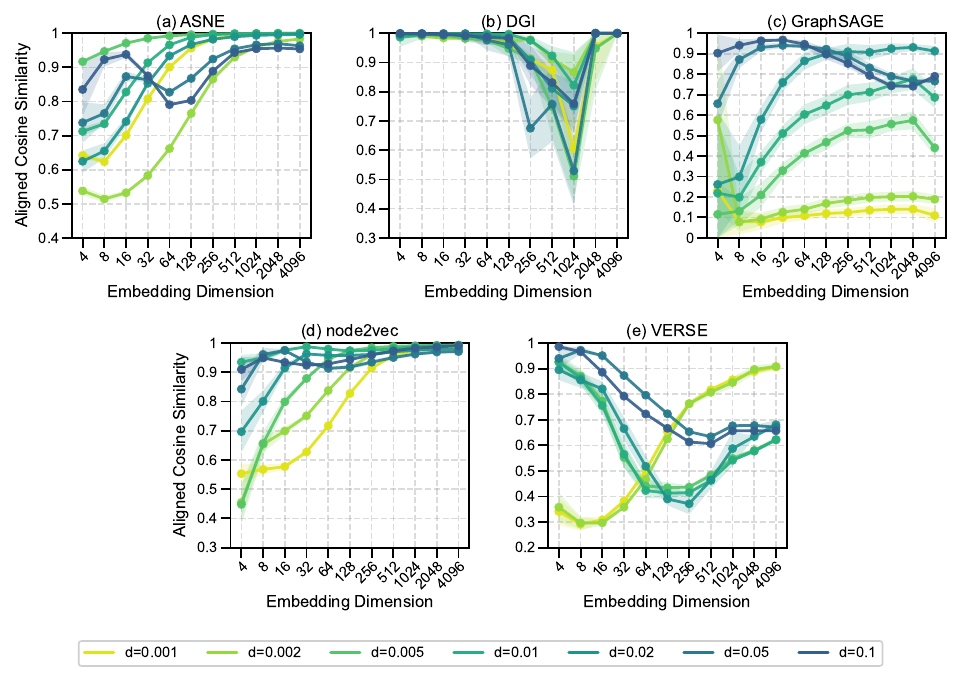}
	\caption{\captionheader{Interplay between network density and embedding dimension on Watts-Strogatz graphs in terms of aligned cosine similarity.}
		We depict aligned cosine similarity averaged across all pairs of 10 embeddings computed on the same graph, aggregated over 10 independently generated Watts-Strogatz graphs with $N=1600$ nodes and varying density as indicated by line color. \new{Shaded bands indicate one standard deviation around the respective means.} 
		Higher values indicate more stable embeddings.
	}
	\label{fig:ws_density_aligned_cossim}
\end{figure}

\begin{figure}[t]
	\centering
	\includegraphics[width=\textwidth]{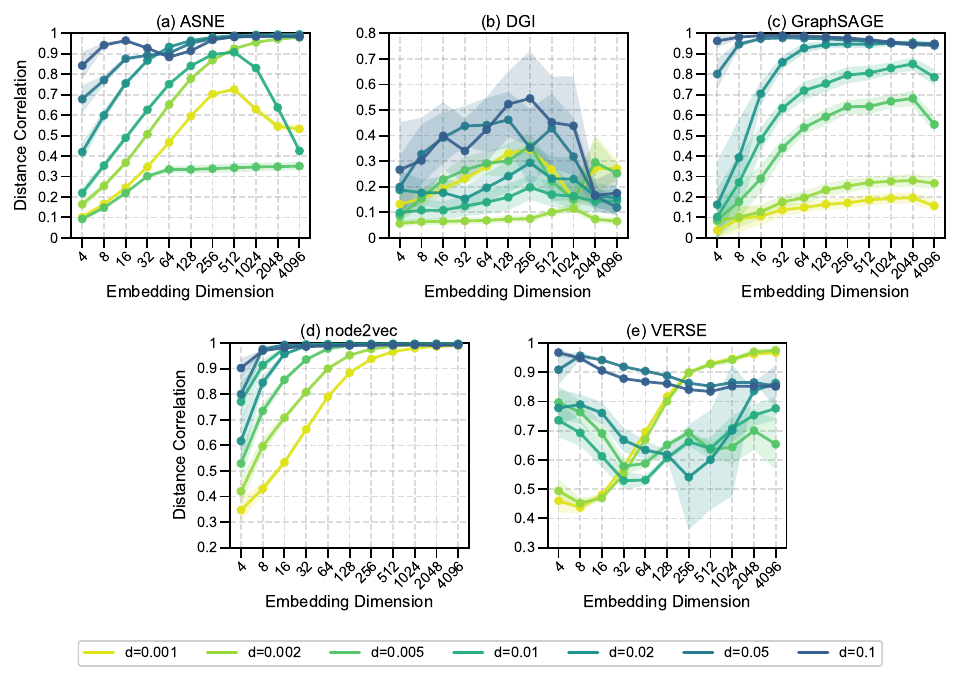}
	\caption{\captionheader{Interplay between network density and embedding dimension on Watts-Strogatz graphs in terms of distance correlation.}
		We depict distance correlation averaged across all pairs of 10 embeddings computed on the same graph, aggregated over 10 independently generated Watts-Strogatz graphs with $N=1600$ nodes and varying density as indicated by line color. \new{Shaded bands indicate one standard deviation around the respective means.} 
		Higher values indicate more stable embeddings.
	}
	\label{fig:ws_density_distcorr}
\end{figure}

Figures \ref{fig:ba_density_jaccard}-\ref{fig:ws_density_distcorr} provide the complete results for the interaction between graph density and embedding dimensionality across both graph models and all representational similarity measures.
Consistent with the discussion in the main text, density-dependent effects are generally less systematic than size-dependent effects.
Although individual methods occasionally exhibit recurring tendencies, these patterns are often weaker and less robust across graph models and similarity measures than those observed for graph size.

\FloatBarrier

\section{Hyperparameter Sensitivity}\label{ap:hyperparameter_sensitivity}

\begin{table}[b]
	\caption{\new{\captionheader{Sensitivity of DGI and GraphSAGE results to dimension-specific hyperparameter tuning on Wikipedia.}
			We compare mean test accuracy and stability scores from the \emph{main} results, based on 30 embeddings using hyperparameters selected at $D=128$, with scores from 10 additional embeddings using hyperparameters \emph{
				tuned} separately at the respective dimension.
			Only dimensions for which the optimal hyperparameters differed from those optimal at $D=128$ are shown.
			For  $k$-NN Jaccard and aligned cosine similarity, higher values indicate greater stability, for disagreement and JSD, this is the case for lower values. Values are rounded to three decimal places.
	}}
	\label{tab:hyperparams}
	\smallskip
	\centering
	{
		\setlength{\tabcolsep}{4pt}
		\footnotesize
		\begin{tabular}{llcccccccccc}
			\toprule
			&  & \multicolumn{2}{c}{\textbf{Accuracy}} & \multicolumn{2}{c}{\makecell[c]{\textbf{k-NN Jaccard}\\ \textbf{Similarity}}} & \multicolumn{2}{c}{\makecell[c]{\textbf{Aligned Cosine}\\ \textbf{Similarity}}} & \multicolumn{2}{c}{\textbf{Disagreement}}  & \multicolumn{2}{c}{\textbf{JSD}}  \\
			\cmidrule(lr){3-4}\cmidrule(lr){5-6}\cmidrule(lr){7-8}\cmidrule(lr){9-10}\cmidrule(lr){11-12}
			& \textbf{Dimension} & \textbf{Main} & \textbf{Tuned} & \textbf{Main} & \textbf{Tuned} & \textbf{Main} & \textbf{Tuned} & \textbf{Main} & \textbf{Tuned} & \textbf{Main} & \textbf{Tuned} \\
			\midrule
			DGI & 4 & 0.273 & 0.219 & 0.013 & 0.009 & 0.517 & 0.503 & 0.655 & 0.582 & 0.001 & 0.001 \\
			\midrule
			GraphSAGE & 2048 & 0.602 & 0.617 & 0.196 & 0.124 & 0.601 & 0.623 & 0.372 & 0.345 & 0.002 & 0.002\\
		 & 4096 & 0.204 & 0.164 & 0.185 & 0.444 & 0.896 & 0.935 & 0.300 & 0.042 & 0.001 & 0.000\\
			\bottomrule
		\end{tabular}
	}
\end{table}

For our main experiments, we determined a set of optimal embedding hyperparameters based on validation performance at dimension $D=128$, and used these parameters to compute embeddings at all other dimensions. 
In the following, we examine the sensitivity of our results to this choice on the Wikipedia dataset, whose size made repeated tuning across dimensions computationally tractable.
We applied the same tuning grid as in the main experiments separately at every dimension.
Whenever the optimal hyperparameters differed from those at dimension $D=128$, we computed ten additional embeddings at this dimension and evaluated their performance and stability.
For downstream tasks, a logistic regressor was used, with the same tuning protocol as in the main experiments.

We find that for ASNE, the resulting hyperparameters do not differ from those determined for $D=128$ at any dimension.
For GraphSAGE, we only observe different optimal parameters at dimensions $D\in\{2048, 4096\}$, while for DGI, parameters differ only at $D=4$.
The corresponding results are reported in \Cref{tab:hyperparams}.
The only notable differences occur for GraphSAGE at dimension 4096: dimension-specific tuning yields lower test accuracy but substantially greater stability, as indicated by higher $k$-NN Jaccard and aligned cosine similarities and lower disagreement and JSD.
Inspection shows that the predicted class probabilities are nearly uniform across classes for most instances, with the most prevalent class---which accounts for 80 of the 482 test instances---receiving the highest probability for 95-97\% of instances by a very slight margin. Thus, the low disagreement primarily reflects largely uninformative majority-class predictions.

\begin{figure}[b]
	\centering
	\includegraphics[width=\textwidth]{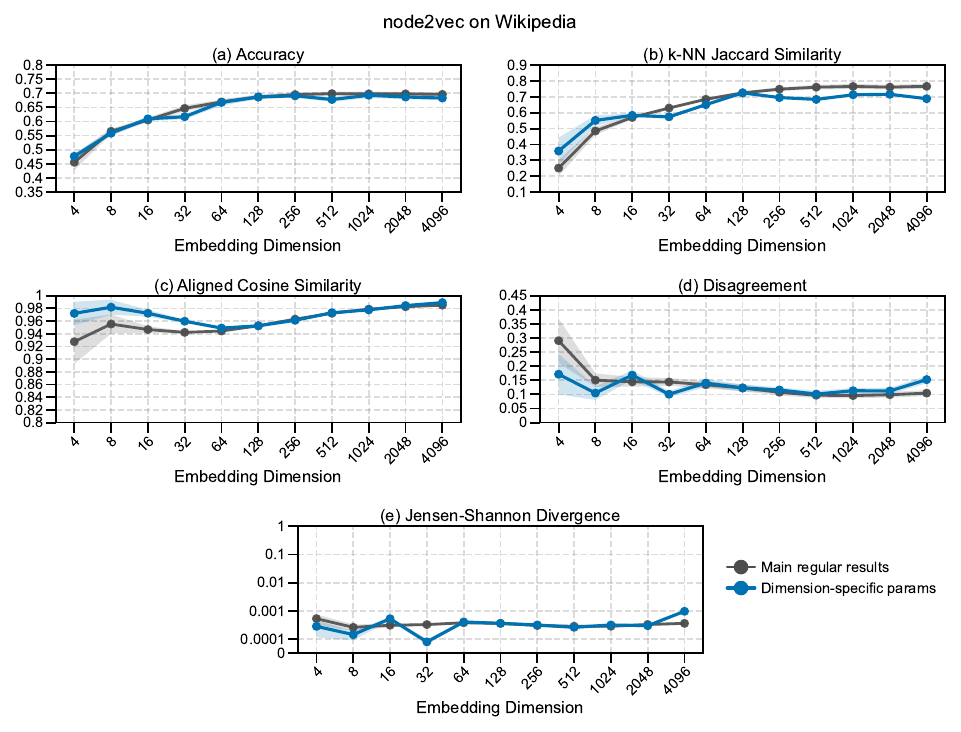}
	\caption{\new{\captionheader{Sensitivity of node2vec results to dimension-specific hyperparameter tuning on Wikipedia.}
			Gray curves show mean performance and stability scores from the 30 embeddings used in the main experiments, where hyperparameters that were selected at $D=128$ are used across dimensions.
			Blue curves show the corresponding means from 10 embeddings using hyperparameters obtained from tuning separately at each dimension.
			Shaded bands indicate one standard deviation around the respective means.
			At $D=128$, both curves show the same result from the main experiments.
	}}
	\label{fig:hyperparams_n2v}
\end{figure}

\begin{figure}[b]
	\centering
	\includegraphics[width=\textwidth]{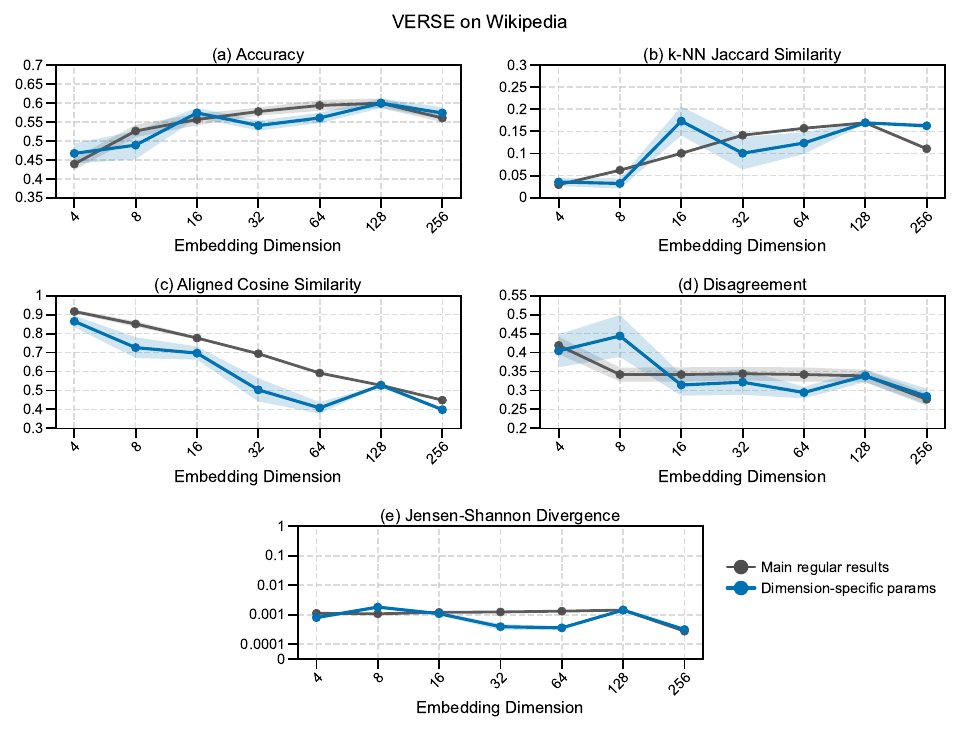}
	\caption{\new{\captionheader{Sensitivity of VERSE results to dimension-specific hyperparameter tuning on Wikipedia.}
			Gray curves show mean performance and stability scores from the 30 embeddings used in the main experiments, where hyperparameters that were selected at $D=128$ are used across dimensions.
			Blue curves show the corresponding means from 10 embeddings using hyperparameters obtained from tuning separately at each dimension.
			Shaded bands indicate one standard deviation around the respective means.
			At $D=128$, both curves show the same result from the main experiments.
	}}
	\label{fig:hyperparams_verse}
\end{figure}

For node2vec, which had the biggest parameter grid, we observe changes in optimal hyperparameters at every dimension different from $D=128$.
Similarly, for VERSE, we also observe different hyperparameters at all dimensions for which tuning could be completed---at dimensions $D>256$, hyperparameter tuning failed at several parameter configurations.
We present the resulting accuracy and stability curves in \Cref{fig:hyperparams_n2v} and \Cref{fig:hyperparams_verse}.
For node2vec, the dimension-specific curves closely follow the main results across performance and stability measures.
VERSE exhibits larger differences at individual dimensions, which are also often associated with higher variance in the reported values. In general, the overall trajectories of accuracy and stability measures appear to be retained.

Overall, dimension-specific tuning neither systematically improves performance or stability nor substantially changes the principal dimensionality-dependent trends observed on Wikipedia.

\FloatBarrier

\section{Computational Costs}\label{ap:costs}

\begin{table}[b!]
	\caption{\new{\captionheader{Embedding and downstream evaluation costs on the Facebook dataset.} We report mean wall-clock time and mean peak RAM use across three consecutive runs on a fixed compute node. Values are rounded to two decimal places.}}
	\label{tab:embedding_costs_facebook}
	\smallskip
	\centering
	\begin{tabular}{llrrrr}
		\toprule
		&  & \multicolumn{2}{c}{\textbf{Embedding Generation}} & \multicolumn{2}{c}{\textbf{Downstream Run}} \\\cmidrule(lr){3-4} \cmidrule(lr){5-6}
		& \textbf{Dimension}  & \textbf{Time (s)} & \textbf{Peak RAM (GB)} & \textbf{Time (s)} & \textbf{Peak RAM (GB)} \\
		\midrule
		\multirow[t]{3}{*}{ASNE} & 16 & 80.57 & 2.10 & 0.33 & 0.01 \\
		& 256 & 108.88 & 2.17 & 0.87 & 0.08 \\
		& 4096 & 222.47 & 3.48 & 111.99 & 1.11 \\
		\midrule
		\multirow[t]{3}{*}{DGI} & 16 & 271.11 & 0.27 & 0.57 & 0.01 \\
		& 256 & 284.24 & 0.18 & 11.94 & 0.03 \\
		& 4096 & 371.28 & 1.08 & 190.83 & 1.11 \\
		\midrule
		\multirow[t]{3}{*}{GraphSAGE} & 16 & 273.66 & 0.04 & 0.34 & 0.02 \\
		& 256 & 399.31 & 0.03 & 1.36 & 0.03 \\
		& 4096 & 3356.39 & 1.10 & 27.06 & 1.12 \\
		\midrule
		\multirow[t]{3}{*}{node2vec} & 16 & 94.80 & 0.33 & 0.32 & 0.01 \\
		& 256 & 262.95 & 0.40 & 1.35 & 0.08 \\
		& 4096 & 3050.00 & 1.43 & 31.42 & 1.11 \\
		\midrule
		\multirow[t]{3}{*}{VERSE} & 16 & 134.71 & 0.02 & 0.26 & 0.00 \\
		& 256 & 190.05 & 0.07 & 0.35 & 0.08 \\
		& 4096 & 2078.39 & 1.10 & 3.26 & 1.11 \\
		\bottomrule
	\end{tabular}
\end{table}

\Cref{tab:embedding_costs_facebook,tab:embedding_costs_coauthor} summarize the computational costs of embedding generation and downstream evaluation at representative dimensions spanning several orders of magnitude on the Facebook and CoAuthor datasets.
These are among the larger datasets evaluated in this study, and represent both the node classification and link prediction downstream tasks, respectively.
These measurements were taken on the \emph{bwForCluster Helix}\footnote{\url{https://wiki.bwhpc.de/e/Helix}}, where also the large majority of the main experiments was conducted.
DGI and GraphSAGE embeddings were computed using a \emph{NVIDIA A100 GPU}; the remaining embeddings were computed on CPUs using \emph{AMD EPYC 7513} processors using 32 workers. Downstream runs in this measurement were executed as single-process CPU jobs.

Under these settings, ASNE exhibited the lowest embedding-generation times in most configurations, with runtimes increasing only moderately across dimensions.
For DGI, runtimes varied more strongly between the two datasets than across dimensions within either dataset.
GraphSAGE, node2vec, and VERSE instead showed marked runtime increases toward higher dimensions. This increase is most pronounced for GraphSAGE, for which we also had to omit computing embeddings for the main experiments at dimensions $D>1024$ on CoAuthor because individual runs exceeded ten hours.

Peak host-RAM requirements generally increased with dimensionality. On CoAuthor, the methods for which dimension 4096 was completed required approximately 32-67\,GB of RAM.
For the GPU-based methods, maximum VRAM consumption (not reported in the tables) reached 77.7\,GB for DGI at dimension 4096 and 26.6\,GB for GraphSAGE at dimension 1024 on CoAuthor. On Facebook, the corresponding maxima ranged between 6 and 8\,GB at dimension 4096.

Regarding downstream tasks, we observe some variation across embedding methods, but this variation remains in the same order of magnitude. 
Notably, node classification was substantially more efficient than link prediction, where the full downstream data was also much larger than for node classification. 
We observe extremely high runtimes of up to 4.5 hours per run on CoAuthor once dimension exceeds 1000.
For that reason, the main experiments ran many such evaluations concurrently, both across CPU cores within individual compute nodes and across multiple nodes.
Nevertheless, on the OGBL-DDI dataset, which is even larger in its downstream data, link prediction evaluation was infeasible for dimensions $D>1024$.

\begin{table}[htbp]
	\caption{\new{\captionheader{Embedding and downstream evaluation costs on the CoAuthor dataset.} We report mean wall-clock time and mean peak memory use across three consecutive runs on a fixed compute node. Values are rounded to two decimal places.}}
	\label{tab:embedding_costs_coauthor}
	\smallskip
	\centering
	\begin{tabular}{llrrrr}
		\toprule
		&  & \multicolumn{2}{c}{\textbf{Embedding Generation}} & \multicolumn{2}{c}{\textbf{Downstream Run}} \\\cmidrule(lr){3-4} \cmidrule(lr){5-6}
		& \textbf{Dimension}  & \textbf{Time (s)} & \textbf{Peak RAM (GB)} & \textbf{Time (s)} & \textbf{Peak RAM (GB)} \\
		\midrule
		\multirow[t]{4}{*}{ASNE} & 16 & 134.13 & 10.34 & 3.68 & 0.16 \\
		& 128 & 157.75 & 10.96 & 4.52 & 1.11 \\
		& 1024 & 168.27 & 15.75 & 5351.23 & 3.33 \\
		& 4096 & 257.00 & 32.35 & 14576.26 & 13.28 \\
		\midrule
		\multirow[t]{4}{*}{DGI} & 16 & 12666.69 & 2.19 & 8.36 & 0.18 \\
		& 128 & 8812.44 & 2.09 & 22.29 & 1.07 \\
		& 1024 & 10516.05 & 16.61 & 6262.16 & 3.34 \\
		& 4096 & 11366.68 & 66.67 & 15555.36 & 13.26 \\
		\midrule
		\multirow[t]{3}{*}{GraphSAGE} & 16 & 1195.99 & 0.28 & 3.40 & 0.16 \\
		& 128 & 1881.59 & 2.09 & 3.78 & 1.05 \\
		& 1024 & 6375.50 & 16.59 & 5483.21 & 3.33 \\
		\midrule
		\multirow[t]{4}{*}{node2vec} & 16 & 221.49 & 0.52 & 3.94 & 0.20 \\
		& 128 & 467.76 & 1.47 & 9.80 & 1.14 \\
		& 1024 & 2953.24 & 8.79 & 5299.18 & 3.36 \\
		& 4096 & 12487.78 & 33.98 & 14983.16 & 13.27 \\
		\midrule
		\multirow[t]{4}{*}{VERSE} & 16 & 338.40 & 0.17 & 3.72 & 0.16 \\
		& 128 & 578.48 & 1.11 & 4.59 & 1.09 \\
		& 1024 & 4712.52 & 8.43 & 5358.42 & 3.33 \\
		& 4096 & 10364.85 & 33.59 & 12480.96 & 13.22 \\
		\bottomrule
	\end{tabular}
\end{table}

\newpage
\FloatBarrier
\section{Optimal Dimensions in Terms of MinGE}\label{ap:min_ge}

To obtain a reference for what would be expected to be good-performing dimensions, we computed the optimal embedding dimensions via the \emph{MinGE} method \citep{luo_graph_2021} for all empirical datasets considered in this study.
This approach is intended to guide dimension selection for the embedding layers of graph neural networks.
The results are reported in \Cref{tab:min_ge}, where we also varied the parameter $\lambda$, which controls the weighting of structural and feature entropy in the full MinGE term. The range of values for $\lambda$ in this table corresponds to the range considered in the hyperparameter study that the authors themselves conducted. Further, $\lambda=1$ is the recommended default value.

Overall, for $\lambda\leq 1$, the optimal dimensions obtained lie almost completely in the interval $(64,128)$, whereas optimal performance in our experiments is usually obtained at dimensions $D\geq256$. Similarly, dimensions of minimum or maximum stability are typically different.
This indicates that MinGE is not well suited to selecting dimensions for the node embedding methods under study.
Further, it is of note that MinGE does not consider raw node attributes in its dimensionality estimations, which are, however, present for all node classification datasets in our study.

\begin{table}[h]
	\centering
	\captionsetup{width=\linewidth}
	\caption{Optimal embedding dimensions obtained from MinGE for different values of the weighting parameter $\lambda$.}
	\label{tab:min_ge}
	\smallskip
	\begin{tabular}{lcccc}
		\toprule
		\textbf{Dataset} & $\lambda=0.1$ & $\lambda=0.5$ & $\lambda=1$ & $\lambda=2$ \\
		\midrule
		Cora        & 69 & 82  & 98  & 130 \\
		PubMed      & 86 & 102 & 123 & 163 \\
		BlogCatalog & 74 & 89  & 106 & 142 \\
		Facebook    & 87 & 103 & 124 & 164 \\
		Wikipedia   & 68 & 80  & 97  & 129 \\
		OGBL-DDI    & 73 & 87  & 104 & 139 \\
		CoAuthor    & 94 & 112 & 135 & 179 \\
		\bottomrule
	\end{tabular}
\end{table}

\end{document}